\documentclass[12pt]{article}
\usepackage{graphicx}
\usepackage{subcaption} 
\usepackage{natbib}
\usepackage{amsmath}
\usepackage{amssymb}
\usepackage{algorithm}
\usepackage{algpseudocode}
\usepackage{multirow}
\usepackage{enumitem}
\usepackage[commandnameprefix=always]{changes}
\usepackage{tikz} 
\usepackage{xcolor}
\usepackage{amsthm}
\usepackage{acro}
\newtheorem{proposition}{Proposition}
\usetikzlibrary{arrows.meta,positioning,shapes.geometric}

\newcommand{\meanSE}[2]{#1{\scriptsize\,(\,#2\,)}}
\newcommand{\best}[1]{\underline{\textbf{#1}}}
\newcommand{\pdrift}{\pi_{\mathrm{d}}}
\newcommand{\blind}{0}
\DeclareAcronym{acronym}{
    short = PASS,
    long = Probabilistic Adaptive Sampling Strategy
}

\begin{document}


\if0\blind
{
  \title{\bf An Adaptive Sampling Framework for Detecting Localized Concept Drift under Label Scarcity}
  \author{Junghee Pyeon \\
    H. Milton Stewart School of Industrial \& Systems Engineering \\
    Georgia Institute of Technology\\
    and \\
    Davide Cacciarelli \\
    EDF Trading \\
    and\\
    Kamran Paynabar \\
    H. Milton Stewart School of Industrial \& Systems Engineering \\ 
    Georgia Institute of Technology\\}
  \maketitle
} \fi

\if1\blind
{
  \bigskip
  \bigskip
  \bigskip
  \begin{center}
    {\LARGE\bf An Adaptive Sampling Framework for Detecting Localized Concept Drift under Label Scarcity}
\end{center}
  \medskip
} \fi

\bigskip
\begin{abstract}
\noindent
Concept drift and label scarcity are two critical challenges limiting the robustness of predictive models in dynamic industrial environments. Existing drift detection methods often assume global shifts and rely on dense supervision, making them ill-suited for regression tasks with local drifts and limited labels. This paper proposes an adaptive sampling framework that combines residual-based exploration and exploitation with EWMA monitoring to efficiently detect local concept drift under labeling budget constraints. Empirical results on synthetic benchmarks and a case study on electricity market demonstrate superior performance in label efficiency and drift detection accuracy.
\end{abstract}

\noindent%
\vfill

\newpage
\section{Introduction}
\label{sec:intro}

Predictive models have become indispensable tools for intelligent decision-making across a wide range of industrial domains, including manufacturing, energy, healthcare, and finance. From demand forecasting to fault detection, these models enable organizations to extract actionable insights from historical and real-time data. Moreover, as industries rely on data-driven decision support systems, the performance and reliability of deployed predictive models play an increasingly essential role in ensuring operational efficiency, safety, and competitiveness. Most of these predictive models are supervised models whose objective is to learn a function $ f: \mathcal{X} \rightarrow \mathcal{Y} $ that maps input features $ \mathbf{x} \in \mathcal{X} $ to a target response $ y \in \mathcal{Y} $, based on a finite collection of labeled examples $ \{(\mathbf{x}_i, y_i)\}_{i=1}^n $. Once trained, this function is used to make predictions on new, unseen data. In theory, this process assumes that both the training and future data are drawn from the same underlying distribution or that a sufficient number of labeled instances are available to capture meaningful patterns.

In practice, however, these assumptions are often violated in dynamic industrial environments. Predictive models frequently encounter two major challenges: (i) nonstationarity of the data distribution over time, and (ii) limited access to labeled data. Nonstationarity, or distribution shift, can occur in various forms \citep{Gama2004,Lu2018}. A common and especially challenging form is concept drift, where the conditional distribution $ P(y \mid \mathbf{x}) $ changes over time \citep{Soares2015,Baier2021,quintana}. This type of drift alters the underlying relationship between inputs and outputs, directly degrading the predictive accuracy of the model. Unlike shifts in the marginal distribution $ P(\mathbf{x}) $ (often referred to as covariate shift), concept drift affects the target-generating mechanism itself, requiring not only detection, but also model adaptation or retraining to restore predictive reliability. Concept drift can manifest in different ways. In the presence of a global drift, the change in the relationship between $ \mathbf{x} $ and $ y $ affects the entire input space uniformly, as might occur after a major system reconfiguration. In contrast, a localized concept drift arises when the change is confined to specific subregions of the input space. For example, a fault condition affecting only a particular machine or a behavioral shift in a subset of customers. Detecting such local changes requires not only temporal sensitivity but also spatial awareness in how drift unfolds across different regions of the input space. Simultaneously, in many real-world applications we need to deal with label scarcity. While input features $ \mathbf{x} $ may be collected continuously via sensors or automated logs, acquiring corresponding output labels $ y $ is often expensive or delayed. In fact, in many cases obtaining labels requires manual inspection, costly experiments, or intrusive treatments.

Together, concept drift and label scarcity severely limit the robustness and adaptability of predictive models in industrial applications. These two challenges have motivated a growing body of research at the intersection of active learning and drift detection, aiming to identify and label only the most informative data points in order to maintain accuracy while minimizing supervision costs. Indeed, several studies have proposed frameworks to combine active learning with concept drift adaptation \citep{Krawczyk2017,Mohamad2018,Shan2019,Zhang2018,Liu2021,sun2018concept,OALSurvey}. However, the vast majority of these approaches have been designed for classification tasks and implicitly assume the presence of a global drift, i.e., a drift that affects the model uniformly across the entire input space. In such cases, the primary goal is to identify when a drift has arisen, so the model can focus on collecting data reflective of the new concept and quickly adapt. In contrast, in many real-world scenarios, drift is not global but local, occurring in specific subregions of the input space while leaving the remainder of the model unchanged. Such subpopulation shifts are common in applications where processes operate under multiple regimes, customer behavior differs across segments, or environmental conditions affect only part of the system. In these cases, addressing drift requires both temporal and spatial sensitivity. Specifically, the detection system must determine not only when a drift has occurred, but also where in the input space the relationship has changed. 

To detect and diagnose concept drifts, recent research has proposed a method that tracks the Fisher score vector—the gradient of the log-likelihood with respect to model parameters, under the assumption that its expectation remains zero under stationarity \citep{zhang2023concept}. This method offers several advantages over traditional error-based drift detection: it is sensitive to internal parameter changes even when residual errors remain stable, it generalizes to broad model classes, and it provides intrinsic diagnostic capabilities. This work marks a considerable step towards regression-based drift detection, but still relies on full data observation or uniform sampling, thus not focusing on the case of localized concept drifts with label scarcity.

In parallel, another strand of literature has explored adaptive sampling strategies for online monitoring under partial observability. A foundational contribution by \citet{liu2015adaptive} introduced the TRAS algorithm, which selects the most informative data streams under resource constraints by ranking local CUSUM statistics. Later works extended this idea to leverage structural correlations among variables. For instance, \citet{nabhan2021correlation} proposed a dynamic sampling algorithm using correlation structures and confidence bounds to infer values for unobserved streams, improving efficiency in change detection. In high-dimensional settings where data arrive as partially observed streams, \citet{xian2021rank} proposed a rank-based sampling algorithm with data augmentation, allowing fast inference of global shifts using only a subset of variables. This idea was extended by \citet{zan2023spatial}, who introduced a spatial rank-based method for nonparametric monitoring and equitable sampling of unobservable but correlated streams. A different but related line of work by \citet{estradagomez2022adaptive} introduced a low-rank tensor recovery framework that adaptively selects sampling locations while capturing the latent structure of high-dimensional data. Similarly, \citet{reisi2019adaptive} tackled data fusion challenges, proposing an adaptive strategy for acquiring high-accuracy labels based on surrogate models built from low-accuracy data.

Despite these advances, most existing methods fall short in addressing local concept drift in regression models under label scarcity. Techniques tailored to mean-shift detection do not account for the more subtle, model-internal changes associated with drift. Many sampling strategies lack principled diagnostics for prioritizing data from spatially drifting regions. And Score vector monitoring, while theoretically sound, has yet to be integrated with adaptive, label-efficient exploration strategies. In this work, we propose \ac{acronym}, which is a novel adaptive sampling strategy for regression models experiencing local concept drift by developing a residual-based sampling framework that allows for efficient and targeted exploration of the input space. 

Specifically, we make the following contributions: 1) we propose \ac{acronym}, a residual-informed adaptive sampling strategy that exploits both the magnitude of the residual and the variability of predictions to focus on regions where drift is most likely to occur; and 2) we integrate this sampling strategy with exponentially weighted moving averages (EWMA), enabling sensitive detection of both abrupt and incremental drifts. The proposed framework is validated on both synthetic benchmarks and a real-world dataset from the electricity market, showing superior performance in detecting local drift while requiring fewer labeled samples.

Our framework sits at the intersection of adaptive sampling, statistical process monitoring, and concept drift detection. It advances the state of the art by targeting the challenging regime of supervised regression modeling under local drift and partial labeling. Although we focus on regression in this paper, the design is model-agnostic and extends naturally to classification settings. The remainder of this paper is organized as follows. Section~\ref{sec:meth} introduces the \ac{acronym} methodology, including both exploration and exploitation strategies as well as the residual-based EWMA monitoring approach. Section~\ref{sec:sim} presents a comprehensive simulation study designed to evaluate the method’s ability to detect localized concept drift under various scenarios. In Section~\ref{sec:case}, we demonstrate the practical utility of the proposed method through a real-world case study based on electricity price monitoring in the UK market. Finally, Section~\ref{sec:conc} concludes the paper with a discussion of implications, limitations, and directions for future work.

\section{Methodology}
\label{sec:meth}

Detecting localized concept drift and responding to it in a timely manner are crucial for maintaining robust predictive performance, particularly in dynamic environments where the conditional distribution $ P(Y \mid X) $ may change over time within specific subregions of the input space but not across the entire domain space. Unlike global drift, which affects the entire input space uniformly, localized drift occurs only in specific regions, making detection subtler and more challenging. Since the input-output relationship changes only within certain subregions of the domain, identifying localized drift using traditional approaches, which assume a uniform change, is problematic. In such scenarios, a detection methodology must be both temporally sensitive (to detect \emph{when} drift occurs) and spatially discriminative (to detect \emph{where} it occurs).

\subsection{Problem Statement}
Suppose a regression or classification model, denoted by $\hat f$ was trained using a set of historical streaming data.  In the online phase, at each time step $ t $, a large batch of unlabeled input observations $ \{\mathbf{x}_{t,1}, \dots, \mathbf{x}_{t,N}\} \subset \mathcal{X} $ becomes available for inspection. The corresponding labels $ \{y_{t,1}, \dots, y_{t,N}\} \subset \mathcal{Y} $, however, are not observed unless explicitly requested. Due to resource constraints, only a limited number of labels can be acquired at each time step. The objective is to design a sampling and monitoring strategy that efficiently detects changes in the conditional distribution $ P(y \mid \mathbf{x}) $, i.e., concept drift, using only the limited labeled feedback available at each time. Due to these constraints, effective drift detection requires strategic allocation of the labeling budget to the most informative parts of the input space. This involves balancing two competing objectives:
\begin{itemize}
    \item Exploitation: Allocating labels to regions that already exhibit signals of instability or elevated residuals in order to confirm and localize potential drift.
    \item Exploration: Allocating labels to sparsely sampled or previously stable regions, to detect emerging or unexpected drifts.
\end{itemize}

\begin{table}[!htb]
\centering
\footnotesize
\setlength{\tabcolsep}{4pt}
\renewcommand{\arraystretch}{1.2}
\caption{Preliminary notations}
\label{tab:prelim-notations}
\begin{tabular}{@{}l l@{}}
\hline
\textbf{Symbol} & \textbf{Meaning} \\
\hline
$t\in\mathbb{N}$ & Time index \\
$\mathcal{D}_t$ & Labeled data up to $t$ \\
$\hat f:\mathcal{X}\to\mathcal{Y}$ & Predictive model \\
$\epsilon\in[0,1]$ & Exploration share \\
$M\in\mathbb{N}$ & Budget per time step \\
$m_x, m_e$ & Budgets for exploitation and exploration \\
$\mathcal{S}_{t,x}, \mathcal{S}_{t,e}$ & Sets of exploitation and exploration selections at time $t$\\
$e_t$ & Residuals from newly labeled data at time $t$ \\
$Z_t^{(\cdot)}$ & EWMA statistic \\
$UCL^{(\cdot)}$ & Corresponding upper control limit \\
\hline
\end{tabular}
\end{table}

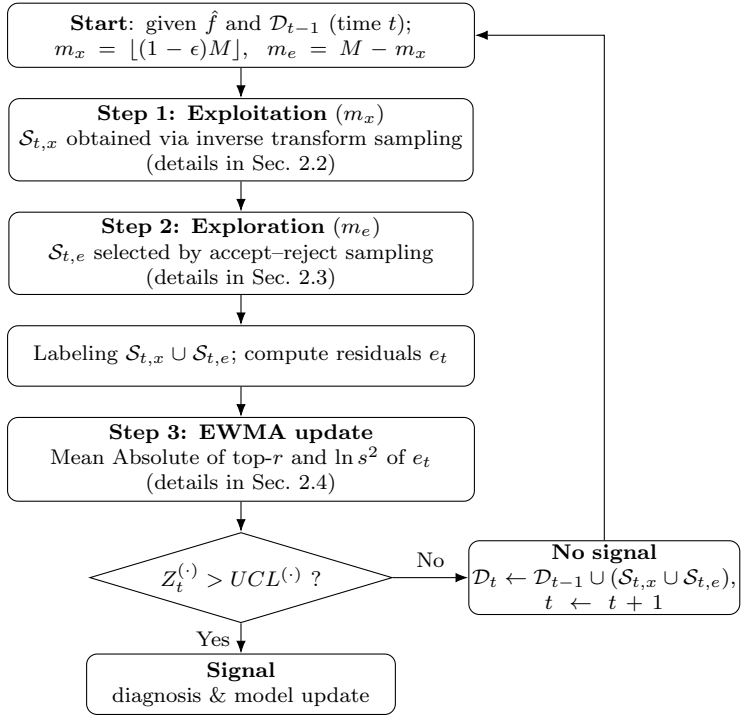
\begin{figure}[!htb]
\centering
\resizebox{0.6\linewidth}{!}{%
\begin{tikzpicture}[
  node distance=4mm and 10mm, >=Latex,
  every node/.style={font=\scriptsize},
  block/.style={rectangle,draw,rounded corners,align=center,inner sep=2.5pt, text width=60mm, minimum height=8mm},
  small/.style={rectangle,draw,rounded corners,align=center,inner sep=2.0pt, text width=60mm, minimum height=8mm},
  decision/.style={diamond,draw,aspect=3,align=center,inner sep=1.0pt, minimum width=40mm},
  mini/.style={rectangle,draw,rounded corners,align=center,inner sep=1.2pt, text width=40mm, minimum height=8mm},
  mini2/.style={rectangle,draw,rounded corners,align=center,inner sep=1.2pt, text width=35mm, minimum height=8mm}
]
\node[block] (start) {\textbf{Start}: given $\hat f$ and $\mathcal{D}_{t-1}$ (time $t$);
$m_x=\lfloor(1-\epsilon)M\rfloor,\ \ m_e=M-m_x$};

\node[small,below=of start] (exploit) {\textbf{Step 1: Exploitation} ($m_x$)\\
$\mathcal{S}_{t,x}$ obtained via inverse transform sampling\\
(details in Sec.~\ref{subsec:exploitation})};

\node[small,below=of exploit] (explore) {\textbf{Step 2: Exploration} ($m_e$)\\
$\mathcal{S}_{t,e}$ selected by accept--reject sampling\\ (details in Sec.~\ref{subsec:exploration})};

\node[block,below=of explore] (label) {Labeling $\mathcal{S}_{t,x} \cup \mathcal{S}_{t,e}$; compute residuals $e_t$};

\node[block,below=of label] (ewma) {\textbf{Step 3: EWMA update} \\ Mean Absolute of top-$r$ and $\ln s^2$ of $e_t$\\
(details in Sec.~\ref{subsec:monitoring})};

\node[decision,below=of ewma] (alarm) {$Z_t^{(\cdot)} > UCL^{(\cdot)}$ ?};
\node[mini,below=of alarm] (signal) {\textbf{Signal} \\ diagnosis \& model update};
\node[mini2,right=of alarm] (update) {\textbf{No signal} \\ $\mathcal{D}_t\leftarrow\mathcal{D}_{t-1}\cup\left( \mathcal{S}_{t,x} \cup \mathcal{S}_{t,e}\right)$, \\ $t\leftarrow t+1$};

\draw[->] (start) -- (exploit);
\draw[->] (exploit) -- (explore);
\draw[->] (explore) -- (label);
\draw[->] (label) -- (ewma);
\draw[->] (ewma) -- (alarm);
\draw[->] (alarm) -- node[left]{Yes} (signal);
\draw[->] (alarm) -- node[above]{No} (update);
\draw[->] (update.north) |- (start.east); 
\end{tikzpicture}%
}
\caption{Overview of the proposed framework; details of each step are described in the corresponding subsections.}
\label{fig:problem_overview}
\end{figure}

An overview of the proposed \ac{acronym} methodology is given in Figure \ref{fig:problem_overview} and Table~\ref{tab:prelim-notations} summarizes the core notations used in the figure and throughout the paper.

As can be seen from the flow diagram, PASS starts with the predictor $\hat f$ and the most recent set of labeled historical data, $\mathcal{D}_{t-1}$. At time $t$, an unlabeled batch (potential observation points in the input space), denoted by $U_t$, arrives. With the budget of $M$ queries per iteration, we split queries into \emph{exploitation} and \emph{exploration}. Exploitation targets regions suspected of drift, whereas exploration ensures coverage of long-unvisited areas; see Sections~\ref{subsec:exploitation}--\ref{subsec:exploration} for details including the non-overlap rule. After querying both exploitation $S_{t,x}$ and exploration $S_{t,e}$ samples, we compute the prediction residuals $e_t$ for labeled samples, and monitor them using one or two one-sided EWMA charts: (i) the mean of the $r$ largest absolute residuals and (ii) the log-variance of residuals. If any monitored statistic exceeds its control limits, we trigger diagnosis or model update; otherwise we append the new labels to $\mathcal{D}$ and proceed to $t{+}1$. The exact implementation is described in Subsection~\ref{subsec:framework}.

To systematically address the balance between exploration and exploitation, \ac{acronym} adopts an $\epsilon$-greedy strategy, widely used by reinforcement learning algorithms \citep{sutton1998reinforcement}, to split budget in each iteration. This strategy provides an intuitive yet powerful approach for managing the exploration-exploitation trade-off by allocating sampling resources probabilistically: with probability $\epsilon$, exploratory sampling is conducted, whereas with probability $1-\epsilon$, the sampling resources are directed toward exploitation. Here, the parameter $\epsilon \in [0, 1]$ explicitly governs the proportion of resources dedicated to exploring unknown or under-examined regions, thus directly influencing the sensitivity and responsiveness of the drift detection procedure. The operational guidelines for tuning $\epsilon$ appears in Subsection~\ref{subsec:param-guidance}.

\subsection{Inverse Transform Sampling for Exploitation}
\label{subsec:exploitation}

Exploitation aims to efficiently leverage previously collected data to enhance drift detection accuracy, particularly under a constrained labeling budget. A common approach in drift detection is to use residuals, the discrepancy between observed and predicted values, as indicators of potential concept drift \citep{Gama2014, Lu2018, krempl2014open}. However, because residuals are observable only for labeled samples, estimating where drift might be occurring across the entire input space becomes inherently challenging. To address this, exploitation in \ac{acronym} uses residual-weighted inverse transform sampling \citep{imberg2020optimal}. This method enables targeted sampling in regions of high residual history, without requiring kernel-based density estimation or exhaustive modeling of the input space.

Given $\mathcal{D}_{t-1}=\{(\mathbf{x}_i,y_i)\}_{i=1}^{n_t}$ and $\hat f$ at time $t$, we compute residuals $e_i=y_i-\hat f(\mathbf{x}_i)$ and non-negative weights $w_i=e_i^2$. Let $\pi_i=w_i/\sum_j w_j$ and define the residual-weighted empirical CDF as $F(k)=\sum_{i\le k}\pi_i$. Inverse transform sampling draws $u\sim\mathrm{Unif}(0,1)$ and selects the anchor index $j=\min\{k: F(k)\ge u\}$; repeating this $m_x$ times yields anchor points without requiring the kernel density estimation \citep{devroye1986sample,robert1999monte}.

Around each selected anchor $\mathbf{x}_j$, we add small Gaussian \emph{turbulence} to generate local candidates at time $t$. We then draw candidates in a spherical neighborhood, and truncate proposals to the valid domain. Sampling exactly at an anchor yields little new information; by probing points that are close but not identical, we retain the anchor’s local behavior while obtaining additional information. 
We therefore use a time-varying scalar perturbation $h_t>0$ to control the diffusion radius around high-residual anchors. Let $\tilde{\mathbf x}_{t,k}$ denote the $k$-th locally perturbed candidate at time $t$ generated around anchor $\mathbf{x}_j$. We then sample
\[
\tilde{\mathbf x}_{t,k} \sim \mathcal N\!\big(\mathbf x_j,\, h_t^2 I_d\big),\qquad k=1,\dots,m_x.
\]
The scalar $h_t$ controls locality. If $h_t$ is too large, residual signals dilute; and if $h_t$ is too small, proposals collapse onto the anchor and information gain is limited.

To choose $h_t$, we initialize $h_0$ to match the smallest practically relevant drift width denoted by $\delta_{\min}$. That is $h_0 \approx \delta_{\min}/2$, when such prior knowledge is available. When such information is unavailable, classical kernel bandwidth heuristics (e.g., Silverman’s rule--of--thumb) give a reasonable default~\citep{silverman2018density}. For stability, we clip $h_t$ to $[h_{\min},h_{\max}]$ with $0<h_{\min}\le h_{\max}$. Set the upper bound relative to the exploration grid by taking $h_{\max}\le g_{\min}$, where $g_{\min}$ is the smallest cell width across axes (see Sec.~\ref{subsec:exploration}). As monitoring proceeds, the exploration stage progressively covers non-suspect regions; once such coverage is deemed sufficient, it is preferable to concentrate effort near suspected drift neighborhoods rather than diffusing widely. We capture this exploitation shift by allowing the perturbation to contract over time. The fixed-perturbation case is achieved by holding $h_t$ constant (i.e., $\rho=1$), whereas gradual focusing is achieved with
\[
h_t \;=\; \max\{\,h_{\min},\, \rho\, h_{t-1}\,\},\qquad \rho\in(0,1).
\]

In summary, exploitation stochastically favors anchors with larger past residuals and then perturbs them within a controlled radius $h_t$. If coordinate scales differ across axes, either normalize the scales across coordinates (e.g., min–max to a common range) or make the perturbation anisotropic by using axis-specific bandwidths $h_{t,j}$ and sampling $\tilde{\mathbf x}_{t,k}\sim\mathcal N(\mathbf x_j,\mathrm{diag}(h_{t,1}^2,\ldots,h_{t,d}^2))$. This concentrates labels where local misspecification is most likely, while still injecting small spatial diversity to avoid sampling the exact same point. Collecting $\tilde{\mathbf{x}}_{t,1},\dots,\tilde{\mathbf{x}}_{t,m_x}$ forms the exploitation set $\mathcal{S}_{t,x}=\{\tilde{\mathbf{x}}_{t,k}\}_{k=1}^{m_x}$ . A concise pseudocode is provided in Algorithm~\ref{alg:exploitation}.

\begin{algorithm}[!htb]
\caption{Exploitation: Residual-weighted Inverse transform}
\label{alg:exploitation}
\begin{algorithmic}[1]
\Require Labeled set $\mathcal{D}_{t-1}=\{(\mathbf{x}_i,y_i)\}_{i=1}^{n_{t-1}}$, model $\hat f$, budget $m_x$, perturbation $h_t$
\Ensure Exploitation set $\mathcal{S}_{t,x}$ of size $m_x$
\State Compute residuals $e_i=y_i-\hat f(\mathbf{x}_i)$ and weights $w_i=e_i^2$
\State \textbf{if } $\sum_i w_i = 0$ \textbf{ then } set $w_i \gets 1$ for all $i$ \textbf{ end if}
\State Set probabilities $\pi_i=w_i/\sum_j w_j$ and CDF $F(k)=\sum_{i\le k}\pi_i$
\State $\mathcal{S}_{t,x}\gets \emptyset$
\For{$k=1$ to $m_x$}
  \State Draw $u\sim\mathrm{Unif}(0,1)$ and find smallest index $j=F^{-1}(u)$
  \State Sample $\tilde{\mathbf{x}} \sim \mathcal{N}(\mathbf{x}_j,\,h_t^2 I_d)$
  \State $\mathcal{S}_{t,x}\gets \mathcal{S}_{t,x}\cup\{\tilde{\mathbf{x}}\}$
\EndFor
\end{algorithmic}
\end{algorithm}

\subsection{Accept--Reject Sampling for Exploration}
\label{subsec:exploration}

Exploration seeks regions that may have changed but have not been queried recently. A representative approach by \citet{liu2015adaptive} updates a cell-wise statistic on a fixed grid so that cells left unobserved for longer accrue larger values and are deterministically prioritized; this is practical for discrete or low-dimensional domains (e.g., image grids). In continuous, high-dimensional inputs, however, “recency” is hard to define without an explicit partitioning, and naive binning incurs a combinatorial blow-up. In addition, many cells may remain empty due to sparsity. Related grid-based strategies in adaptive monitoring and spatial sampling face similar limitations in continuous spaces \citep{estradagomez2022adaptive,zan2023spatial}. These considerations motivate a sparse, history-aware exploration scheme that avoids full-grid storage and tracks recency only where data have actually appeared.

To overcome these practical limitations, \ac{acronym} employs an exploration scheme with accept--reject sampling, a technique that is inherently robust to high dimensionality and widely used in simulation and Bayesian inference \citep{robert1999monte, rubinstein2016simulation}. Accept--reject sampling first generates candidate samples from a simple proposal distribution and then accepts or rejects them based on a probability derived from a target distribution. Formally, for a candidate location with acceptance probability $ p $, a uniform random variable $ u \sim \mathrm{Unif}(0,1) $ is drawn. The candidate is accepted if $ p \geq u $; otherwise, a new candidate is sampled. In practice, our adaptive exploration strategy applies this mechanism to the exploration portion of the labeling budget, defined by the parameter $ \epsilon $.

In order to apply this efficient approach, the input domain is partitioned along each axis, inducing a grid $\mathcal{G}$ whose cells are indexed by $c_j; j=1,\dots, |\mathcal{G}|$. We maintain a \emph{sparse} last-visit map $\mathcal{T}\{\tau_{c_j,t}\}$ only for visited cells, replacing any missing $\tau_{c_j}$ with $0$ when needed. At time $t$, exploration runs after exploitation: we first use the most recent $\mathcal{T}\{\tau_{c_j,t}\}$  to draw $m_e=M-m_x$ exploration samples from $\mathcal{G}$. For each exploration draw, randomly select a candidate cell $c_j$ uniformly from $\mathcal{G}$, compute $\Delta t=t-\tau_{c_j}$, and set the acceptance probability $p=\min\!\Big\{\frac{\Delta t}{\min\{t,\,|\mathcal{G}|\}},\,1\Big\}$. Draw $u\sim \mathrm{Unif}(0,1)$; if $u\le p$, accept $c_j$, sample $\mathbf{x_e}$ uniformly within $c_j$, append it to $\mathcal{S}_{t,e}$, and update $\tau_{c_j}\leftarrow t$; otherwise, reject and resample. Repeat until $|\mathcal{S}_{t,e}|=m_e$. Practical choices of the grid resolution, including anisotropic settings and effect-size matching, are discussed in Subsection~\ref{subsec:param-guidance}.

This strategy offers practical advantages over conventional adaptive methods by implicitly promoting exploration without exhaustive history tracking. The accept--reject rule is \emph{time-weighted}, that is cells unvisited for longer receive larger acceptance probabilities $p$, while recently visited cells are down-weighted. This yields broad coverage without storing per-cell histories across all $|\mathcal{G}|$ cells; instead, we only keep the sparse last-visit times $\mathcal{T}\{\tau_{c_j,t}\}$. The normalization by $\min\{t,|\mathcal{G}|\}$ keeps $p\in[0,1]$ uniformly over time and lets the rule adapt smoothly from cold-start to steady-state regimes. Because candidate cells are drawn uniformly, selection pressure depends on \emph{recency} rather than raw visit counts, preventing a few high-traffic areas from monopolizing the exploration budget. Note that the last-visit map $\mathcal{T}\{\tau_{c_j,t}\}$ remains sparse, i.e., only visited cells are stored, so memory and update costs scale with the number of seen cells. Finally, running exploration after exploitation temporarily deprioritizes just-probed regions, making the two stages complementary: exploitation intensifies sampling near suspected drift, while exploration backfills long-unvisited areas. The full procedure is summarized step by step in Algorithm~\ref{alg:exploration}.

\begin{algorithm}[!htb]
\caption{Exploration at time $t$ (accept--reject with sparse last-visit times)}
\label{alg:exploration}
\begin{algorithmic}[1]
\Require Grid $\mathcal{G}$; sparse last-visit map $\mathcal{T}\{\tau_{c_j,t}\}$ (missing $\tau_{c_j}$ interpreted as $0$); current time $t$; budget $m_e$; exploitation set $\mathcal{S}_{t,x}$
\Ensure Exploration set $\mathcal{S}_{t,e}$ with $|\mathcal{S}_{t,e}|=m_e$
\State $\mathcal{S}_{t,e}\gets\emptyset$
\For{$\mathbf{x}\in\mathcal{S}_{t,x}$}
  \State $c_j\gets \mathsf{cell}(\mathbf{x})$ ;\quad $\tau_{c_j}\gets t$
\EndFor
\While{$|\mathcal{S}_{t,e}|< m_e$}
  \State Draw candidate cell $c_j$ uniformly from $\mathcal{G}$
  \State $\tau_{c_j} \gets$ stored $\tau_{c_j}$ if defined, else $0$;\quad $\Delta t \gets t-\tau_{c_j}$
  \State $p \gets \min\!\Big\{\frac{\Delta t}{\min\{t,\,|\mathcal{G}|\}},\,1\Big\}$;\quad draw $u\sim\mathrm{Unif}(0,1)$
  \If{$u \le p$}
     \State Draw $\mathbf{x_{t,e}}$ uniformly in cell $c_j$
     \State $\mathcal{S}_{t,e}\gets \mathcal{S}_{t,e}\cup\{\mathbf{x_{t,e}}\}$;\quad $\tau_{c_j}\gets t$
  \EndIf
\EndWhile
\end{algorithmic}
\end{algorithm}

\subsection{EWMA Control Chart for Monitoring}
\label{subsec:monitoring}

To detect a concept drift as new data arrives, we adopt a statistical process monitoring approach based on EWMA control charts. EWMA charts are particularly well-suited for our framework due to their sensitivity to small and gradual shifts in monitored statistics \citep{lucas1990exponentially, montgomery2020introduction}. The EWMA statistic of a generic monitored quantity $\theta$ at time $t$ is
\[
    z_t = \lambda\,\theta_t + (1-\lambda)\,z_{t-1},
\]
where $z_0$ denotes the in-control target, and $\lambda\in[0,1]$ is a smoothing parameter determining the emphasis on recent observations. Smaller $\lambda$ values prioritize historical data, whereas larger values increase sensitivity to recent changes. Typical values range between 0.15 and 0.25 \citep{montgomery2020introduction}. Control limits for a two-sided EWMA chart of $\theta$ are
\[
    UCL/LCL = \theta_0 \pm L \cdot \hat{\sigma}_\theta \sqrt{\frac{\lambda}{2-\lambda}\Bigl(1-(1-\lambda)^{2t}\Bigr)}\,,
\]
with $L$ determined based on the desired type-I error. As $ t \rightarrow \infty $, control limits stabilize. 

Practically, the preceding definition guides the tuning of $\lambda$ and $L$ without breaking the EWMA flow: consistent with the typical range $0.15$–$0.25$ for $\lambda$, we estimate the in-control scale $\hat{\sigma}_\theta$ from a stable baseline and then choose $L$ to meet a target in-control false-alarm rate, often expressed via in-control average run length (ARL$_0$) (e.g., 200–370) under the steady-state variance $\hat{\sigma}_\theta^2\,\lambda/(2-\lambda)$ \citep{lucas1990exponentially,montgomery2020introduction}. Alternatively, when the in-control distribution or dependence structure is uncertain or the design is nonstandard, $L$ can be calibrated empirically via Monte Carlo or bootstrap so that the control limits attain the target in-control false-alarm rate, which is equivalent to the desired ARL$_0$.

Within \ac{acronym}, we utilize two monitoring statistics within the EWMA framework. First, we track the top-$r$ absolute-residual mean
\[
A_t^{(r)}=\frac{1}{r}\sum_{j=1}^{r}\lvert e_t\rvert_{(j)},
\]
where $e_t$ is the residual vector at time $t$ and $\lvert e_t\rvert_{(1)}\ge\dots\ge\lvert e_t\rvert_{(r)}$ denote the $r$ largest absolute residuals at time $t$. Under in-control conditions where residuals concentrate near zero, $A_t^{(r)}$ remains close to zero; however, when localized drift occurs, clusters of large $\lvert e_t\rvert$ drive $A_t^{(r)}$ upward. This choice of the monitoring statistic focuses on the $r$ largest $|e|$ highlighting clustered hotspots.

Second, we monitor dispersion via the log-variance statistic $V_t=\ln s_t^2$, where $s_t^2$ is the sample variance of $e_t$ computed from a batch of size $n$ at time $t$ with degrees of freedom $k=n-1$. Because the raw sample variance $s_t^2$ is non-Gaussian, working on the log scale provides an approximation to normality \citep{crowder1992ewma,johnson1995continuous}. The corresponding mean and variance are
\begin{flalign*}
    \mathbb{E}[\ln{s^2}] &= \ln{\hat{\sigma}^2} - \ln{(n-1)} + \psi\!\left(\frac{k}{2}\right) + \ln{2}, \\
    \mathrm{Var}(\ln{s^2}) &= \psi_1\!\left(\frac{k}{2}\right),
\end{flalign*}
where $\psi(\cdot)$ and $\psi_1(\cdot)$ are the digamma and trigamma functions, respectively. When drifted and non-drifted regions are sampled together, the resulting mixture inflates dispersion; monitoring $\log s_t^2$ is a standard and effective way to capture such variance increases under an EWMA design.

In our application, we use an upper one-sided EWMA for both statistics. First, the top-$r$ absolute-residual mean is nonnegative, so deterioration manifests only as increases; decreases are not indicative of adverse drift. Second, for dispersion we are operationally concerned with increases in residual variance, while decreases are typically benign and not actionable. Writing $\theta_t$ for either $A_t^{(r)}$ or $V_t$ and $\theta_0$ for its in-control target, we update
\[
z_t \;=\; \lambda\,\max\{0,\;\theta_t-\theta_0\} \;+\; (1-\lambda)\,z_{t-1},
\qquad z_0=0,\ \ \lambda\in(0,1].
\]
Only upward deviations contribute through the truncation $\max\{\cdot,0\}$; when $\theta_t\le \theta_0$, the chart holds or decays, so decreases do not trigger alarms. This truncated one-sided EWMA form is standard in the one-sided EWMA literature. \citep{lucas1990exponentially,duong2022revisiting} Either monitor can be used alone, or both can run in parallel. Using both generally increases detection power but also inflates the overall false-alarm rate; in that case, the joint UCLs should be calibrated to a target ARL$_0$ (e.g., via simultaneous one-sided EWMAs or mean–variance joint schemes). \citep{lucas1990exponentially,gan1995joint}

\subsection{Integrated Framework }
\label{subsec:framework}

\begin{algorithm}[!htb]
\caption{Integrated adaptive sampling with parallel monitoring}
\label{alg:framework}
\begin{algorithmic}[1]
\Require $\mathcal{G}$; $\mathcal{T}\{\tau_{c_j,t}\}$; $\hat f$; $\mathcal{D}_t$; $M$; $\epsilon$; $h_t$; $\lambda$; $r$; $\theta_0^{(A)},\theta_0^{(V)}$; $UCL^{(A)},UCL^{(V)}$
\While{monitoring is active}
  \State $t\!\gets\!t{+}1$;\; $m_x\!\gets\!\lfloor(1{-}\epsilon)M\rfloor$;\; $m_e\!\gets\!M{-}m_x$
  \State \textbf{Exploitation:} $\mathcal{S}_{t,x}\gets \textsc{Exploitation}(\mathcal{D}_{t-1},\hat f,m_x,h_t)$ \Comment{Alg.~\ref{alg:exploitation}}
  \State \textbf{Exploration:} $\mathcal{S}_{t,e}\gets \textsc{Exploration}(\mathcal{G},\mathcal{T}\{\tau_{c_j,t}\},t,m_e,\mathcal{S}_{t,x})$ \Comment{Alg.~\ref{alg:exploration}}
  \State Obtain labels for $\mathcal{S}_{t,x}\cup\mathcal{S}_{t,e}$; form residual vector $e_t$
  \State Compute $A_t^{(r)}=\frac{1}{r}\sum_{k=1}^{r}|e_t|_{(k)}$ and $V_t=\ln s_t^2$
  \State \textbf{Update one-sided EWMAs} (truncated to upward deviations):
        \[
        \begin{aligned}
        Z_t^{(A)} &\gets \lambda\,\max\!\{0,\,A_t^{(r)}-\theta_0^{(A)}\} + (1-\lambda)\,Z_{t-1}^{(A)},\\
        Z_t^{(V)} &\gets \lambda\,\max\!\{0,\,V_t-\theta_0^{(V)}\} + (1-\lambda)\,Z_{t-1}^{(V)}.
        \end{aligned}
        \]
  \If{$Z_t^{(A)} > UCL^{(A)}$ \textbf{ or } $Z_t^{(V)} > UCL^{(V)}$}
     \State Signal concept drift;\; trigger diagnosis and model update
  \Else
     \State $\mathcal{D}_t \gets \mathcal{D}_{t-1}\cup \{(\mathbf{x},y):\,\mathbf{x}\in \mathcal{S}_{t,x}\cup \mathcal{S}_{t,e}\}$
  \EndIf
\EndWhile
\end{algorithmic}
\end{algorithm}

We operationalize the method by coupling the two sampling routines with the monitoring layer in a single loop. The pseudocode in Algorithm~\ref{alg:framework} shows the parallel two-chart variant, i.e., top-$r$ and log-variance used together; if a single monitor is preferred, simply drop the other chart’s lines while keeping the same loop. At each time $t$, the labeling budget $M$ is split into exploitation and exploration, $m_x=\lfloor(1-\epsilon)M\rfloor$ and $m_e=M-m_x$. \emph{Exploitation} calls the residual-weighted inverse transform routine (Algorithm~\ref{alg:exploitation}) with turbulence scale $h$ to produce $\mathcal{S}_{t,x}$. \emph{Exploration} then uses the accept--reject procedure with sparse last-visit times (Algorithm~\ref{alg:exploration}), updating timestamps using $\mathcal{S}_{t,x}$ and drawing $\mathcal{S}_{t,e}$. Newly labeled samples yield residuals $e_t$, from which we compute the chosen monitoring statistic(s) $\theta_t\in\{A_t^{(r)},\,V_t\}$ and update upper one-sided EWMAs. If any chart exceeds its calibrated $UCL^{(\cdot)}$, we trigger a drift alarm and proceed to diagnosis/update; otherwise we append the new labels to the historical set and continue. Either monitor can be used alone, or both in parallel with $UCL^{(\cdot)}$ calibrated accordingly.

\subsection{Practical Guidance for Parameter Setting}
\label{subsec:param-guidance}

This subsection provides practitioner guidance for selecting the main settings and tuning parameters when applying \ac{acronym}. Parameters should be tuned jointly rather than in isolation, since $\epsilon$, $B$, $h$, and the EWMA settings interact.

\begin{itemize}[leftmargin=*,itemsep=2pt]
    \item \textbf{$\epsilon$ (share of exploration).}
    We define $\epsilon\in(0,1)$ as the fraction of the labeling budget allocated to exploration; the remaining $1-\epsilon$ is used for residual-driven exploitation. Based on empirical evidence, we recommend staying below heavy-exploration levels (e.g., $\epsilon\approx0.8$), typically around $0.2$–$0.5$, with modest deviations as warranted by application needs. For detection of weak and localized drifts or early confirmation near suspicious areas, select values toward the lower end of this range to secure repeated measurements via exploitation. To address strong or more widespread drifts, or when residual cues are sparse, lean toward the upper end to broaden coverage. In high-dimensional inputs, a very large $\epsilon$ is often ineffective because it diffuses the exploration budget across many candidate regions, lowering revisit probabilities and delaying stabilization of residual evidence; hence a moderate exploration level is generally preferred.
    
    \item \textbf{$\mathcal{G}$ (grid resolution for exploration).}
    Partition each axis into $B_j$ bins, yielding a grid of $|\mathcal{G}|$ cells for the accept--reject revisit logic. Coarse grids tend to mix affected and unaffected areas, whereas very fine grids leaves many cells rarely visited. Let $\delta_{\min}$ denote the smallest drift width of practical interest, we suggest to match the grid resolution to the effect size by choosing a cell width $w \lesssim \delta_{\min_j}/2$, i.e., $B_j \gtrsim 2/\delta_{\min_j}$ .  This resolution–to–effect-size principle accords with spatial scan practice, where the window scale is tuned to the anticipated cluster size to avoid dilution (too coarse) or sparsity (too fine) \citep{kulldorff1997spatial,tango2012flexible}.

    \item \textbf{$r$ (top-$r$ in the residual monitor).}
    In our setting, $A_t^{(r)}$ averages the $r$ largest absolute residuals among the $M$ labels at time $t$, so $1\le r\le M$. To choose $r$, note that \emph{too small $r$} can overreact to isolated spikes or outliers and miss diffuse changes, whereas \emph{too large $r$} dilutes the signal by averaging many near-zero residuals, effectively approaching a plain mean. We recommend linking $r$ to the expected number of affected labels per batch: when only a few points are likely affected, keep $r$ small relative to $M$; when a larger fraction is expected, increase $r$ moderately but keep $r<M$ to avoid dilution. In practice, start small, adjust with a short pilot, and calibrate the UCL to maintain the target ARL$_0$. This aligns with top-$r$ monitoring guidance that ties $r$ to the anticipated number of affected streams or units \citep{mei2011quickest,liu2015adaptive}.

\end{itemize}

\subsection{Theoretical Properties}
\label{subsec:Theory}
 
We now state two properties that justify the design choices above. Proposition~\ref{prop:exploitation} quantifies the chance that exploitation hits a localized drift, while Proposition~\ref{prop:exploration} shows that the exploration rule cannot permanently neglect any cell. Proofs are given in \ref{app:proof_exploitation} and \ref{app:proof_exploration}, respectively.

\begin{proposition}
\label{prop:exploitation}
Let $\mathcal{R}\subset\mathbb{R}^d$ be a drift region with nonempty interior and fix $h>0$. For any $\mathbf{x}\in\mathcal{R}$, if $\tilde{\mathbf{x}}\sim\mathcal{N}(\mathbf{x},h^2 I_d)$, then $\mathbb{P}(\tilde{\mathbf{x}}\in\mathcal{R})>0$. Moreover, given points $\{\mathbf{x}_i\}_{i=1}^n\subset\mathbb{R}^d$, assume an index $I\in\{1,\dots,n\}$ satisfies $\mathbb{P}(I=i)=\pi_i$ with $\pi_i\ge0$ and $\sum_{i=1}^n\pi_i=1$, and, conditional on $I=i$, $\tilde{\mathbf{x}}\mid (I=i)\sim\mathcal{N}(\mathbf{x}_i,h^2 I_d)$. For each $i$, define $r_i:=\inf_{\mathbf{y}\in\mathcal{R}^c} || \mathbf{x}_i-\mathbf{y}||_2$. Then the following lower bound holds:
\[
\mathbb{P}\big(\tilde{\mathbf{x}}\in\mathcal{R}\big)\ \ge\ \sum_{i=1}^n \pi_i\,\mathbb{P}\!\big(\|\mathbf{Z}\|_2\le r_i/h\big),
\quad \text{where }\mathbf{Z}\sim\mathcal{N}(\mathbf{0},I_d).
\]
\end{proposition}

Proposition~\ref{prop:exploitation} implies that once at least one anchor lies inside the drifted region $\mathcal{R}$, a single exploitation proposal hits $\mathcal{R}$ with strictly positive probability and admits the following lower bound. Defining $p_{\mathrm{exp}}:=\sum_{i=1}^n \pi_i\,\mathbb{P}\!\big(\|\mathbf{Z}\|_2\le r_i/h\big)$, we have, for a batch of $m_x\ge1$ independent exploitation proposals,
\[
\mathbb{P}\Big(\exists\,k\le m_x:\ \tilde{\mathbf{x}}^{(k)}\in\mathcal{R}\Big)\ \ge\ 1-\big(1-p_{\mathrm{exp}}\big)^{m_x}.
\]
As additional anchors enter $\mathcal{R}$ or their selection weights increase, the interior mass $\sum_{i:\,\mathbf{x}_i\in\mathcal{R}}\pi_i$ grows, which boosts $p_{\mathrm{exp}}$ and thus raises the chance that exploitation places proposals in $\mathcal{R}$. This concentrates labels around the drifted region and, in turn, amplifies the monitoring summaries, increasing the likelihood of signaling under fixed UCLs.

\begin{proposition}
\label{prop:exploration}
Let $\mathcal{G}$ be a finite input grid. For each cell $c\in\mathcal{G}$, let $\tau_c(t)$ denote its last-visit time up to $t$ and define
\[
p_c(t)\;=\;\min\!\Big\{\frac{t-\tau_c(t)}{\min\{t,\,|\mathcal{G}|\}},\,1\Big\}.
\]
Let $\mathcal{U}$ denote the set of cells that are never visited after a finite time $t_0$, then $\mathbb{P}\!\left(\mathcal{U}=\varnothing\right)=1$.
\end{proposition}

Proposition~\ref{prop:exploration} shows that the exploration rule is intrinsically self-correcting. 
First, the monotone acceptance $p_c(t)=\min\!\{\frac{t-\tau_c(t)}{\min\{t,|\mathcal{G}|\}},1\}$ implies that any cell left unvisited sufficiently long will be accepted upon proposal once $t\ge \tau_c(t)+|\mathcal{G}|$. Second, i.i.d. uniform proposals ensure that every cell is proposed infinitely often. Therefore, no region can be permanently neglected. 
Operationally, whenever a cell is just sampled by either exploitation or exploration framework, its last-visit time updates to the current step so that $\Delta t=t-\tau_c(t)=0$ and thus $p_c(t)=0$ at time $t$, which diverts the next exploration proposals toward other regions until their $\Delta t$ grows. This mechanism yields broad coverage without additional tuning.

\section{Simulation Study}
\label{sec:sim}

In this section, we evaluate the effectiveness of the proposed adaptive sampling framework for localized concept drift detection through a series of controlled simulation experiments.

\subsection{Experimental Design}
\label{sub:design}

By introducing synthetic drifts into benchmark functions with known analytical structure, we assess how well the proposed method balances labeling efficiency and detection accuracy under varying conditions. Specifically, we simulate localized concept drift by perturbing small subregions of the input space in otherwise in-control functions. This controlled setting allows us to isolate and systematically vary key factors such as drift magnitude, spatial extent, and input dimensionality, while comparing the proposed method against baseline and reference techniques.

To ensure a diverse evaluation across different input dimensions and functional forms, we selected four well-known test functions:

\begin{enumerate}
    \item Branin Function~\citep{richter2020model}:
    \begin{equation}
        f(\mathbf{x}) = \left(x_2 - \frac{5.1}{4\pi^2}x_1^2 + \frac{5}{\pi}x_1 - 6\right)^2 + 10\left(1 - \frac{1}{8\pi}\right)\cos(x_1) + 10 + \eta, \quad \eta \sim \mathcal{N}(0,\,11.32^2)
    \end{equation}

    \item Ishigami Function~\citep{ishigami1990importance}:
    \begin{equation}
        f(\mathbf{x}) = \sin(x_1) + 7\sin^2(x_2) + 0.1\,x_3^4\sin(x_1) + \eta, \quad \eta \sim \mathcal{N}(0,\,0.187^2)
    \end{equation}

    \item Friedman Function~\citep{friedman1983multidimensional}:
    \begin{equation}
        f(\mathbf{x}) = 10\sin(\pi x_1 x_2) + 20(x_3 - 0.5)^2 + 10x_4 + 5x_5 + \eta, \quad \eta \sim \mathcal{N}(0,\,0.05^2)
    \end{equation}

    \item Linkletter Function~\citep{linkletter2006variable}:
    \begin{equation}
        f(\mathbf{x}) = \sum_{n=1}^{8} \frac{0.2}{2^{n-1}}x_n + \eta, \quad \eta \sim \mathcal{N}(0,\,1^2)
    \end{equation}
\end{enumerate}

This selection spans a range of dimensionalities, from low (2D) to moderately high (8D), and includes both nonlinear and additive structures. This diversity enables a comprehensive evaluation of the proposed adaptive sampling framework under varying levels of complexity and input sparsity. 

To simulate localized drift, we inject a shift on an axis-aligned hypercube across all coordinates. The center components are drawn uniformly per coordinate: $x_j^\ast\sim\mathrm{Unif}[LB_j+w_j,\,UB_j-w_j]$ for $j=1,\ldots,d$, ensuring full containment where $LB_j$ and $UB_j$ denote the lower and upper bounds of coordinate $j$, respectively. For a target affected\mbox{-}volume ratio $\pdrift$ over the full domain $\prod_{j=1}^d[LB_j,UB_j]$, we set the half\mbox{-}widths
\[
w_j \;=\; \tfrac12\,\pdrift^{1/d}\,(UB_j-LB_j), \qquad j=1,\ldots,d.
\]
We define the drift region $\mathcal R := \{\mathbf x:\ |x_j-x_j^\ast|\le w_j\ \ \forall j=1,\ldots,d\}$ and set
\[
y \;=\; f(\mathbf x)+\eta \;+\; \Delta\times\mathbf 1\!\{\mathbf x\in \mathcal R\}.
\]
Here, $\Delta$ controls the severity of the drift and is reported in multiples of the noise standard deviation $\sigma$ for comparability across functions, while $\pdrift$ denotes the affected-volume fraction, i.e., $\pdrift = |\mathcal R| \big/ \prod_{j=1}^d(UB_j-LB_j)$.

Each simulation used a per-step sampling budget of $M=20$, with exploitation (residual-weighted inverse transform) and exploration (time-weighted accept--reject) executed in that order. For exploitation, we set a constant perturbation $h_t\equiv h$ (i.e., $\rho=1$), choosing $h$ to be effective when the localized drift occupies at least $1\%$ of the active subspace. The exploration candidate grid uses $B=20$ bins per axis for the Branin, $B=10$ for Ishigami, $B=6$ for Friedman, and $B=4$ for Linkletter, yielding $B^d$ candidates per step in each case.

For all functions except Linkletter, the predictive model is a spline-with-interactions regression. Specifically, B-spline bases are constructed per coordinate, and then expanded by pairwise interaction features (degree-2, interaction-only) before fitting a Ridge regression model; for the Linkletter function we use ordinary least squares. Monitoring employs two one-sided EWMAs on residual summaries (mean of top-$r$ absolute residuals and $\ln s^2$), with smoothing parameter $\lambda=0.2$.

\begin{figure}[!htb]
  \centering
  \begin{subfigure}[t]{0.32\linewidth}
    \centering
    \includegraphics[width=\linewidth]{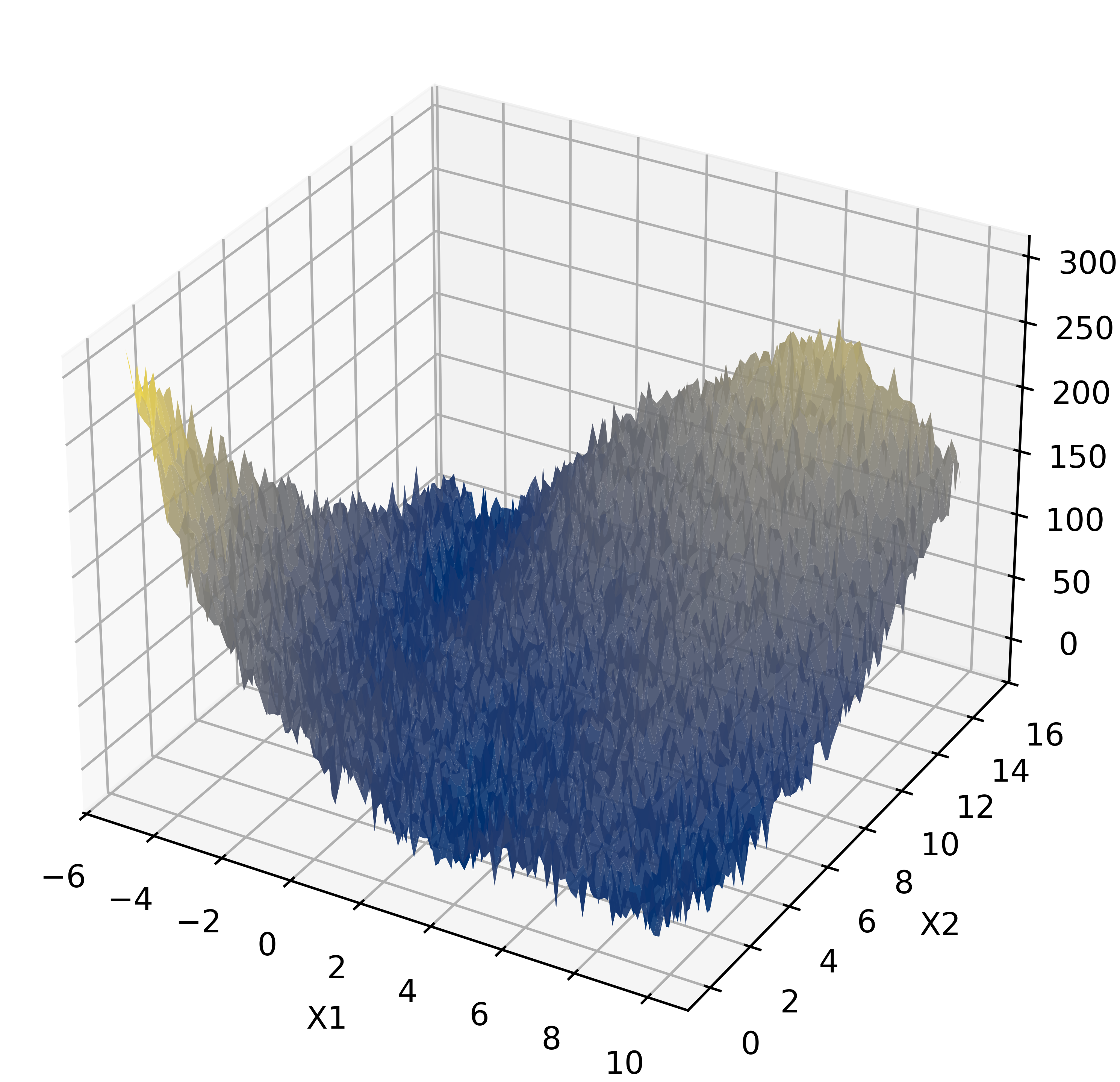}
    \caption{Initial dataset $\mathcal{D}_0$}
    \label{fig:function_init}
  \end{subfigure}\hfill
  \begin{subfigure}[t]{0.32\linewidth}
    \centering
    \includegraphics[width=\linewidth]{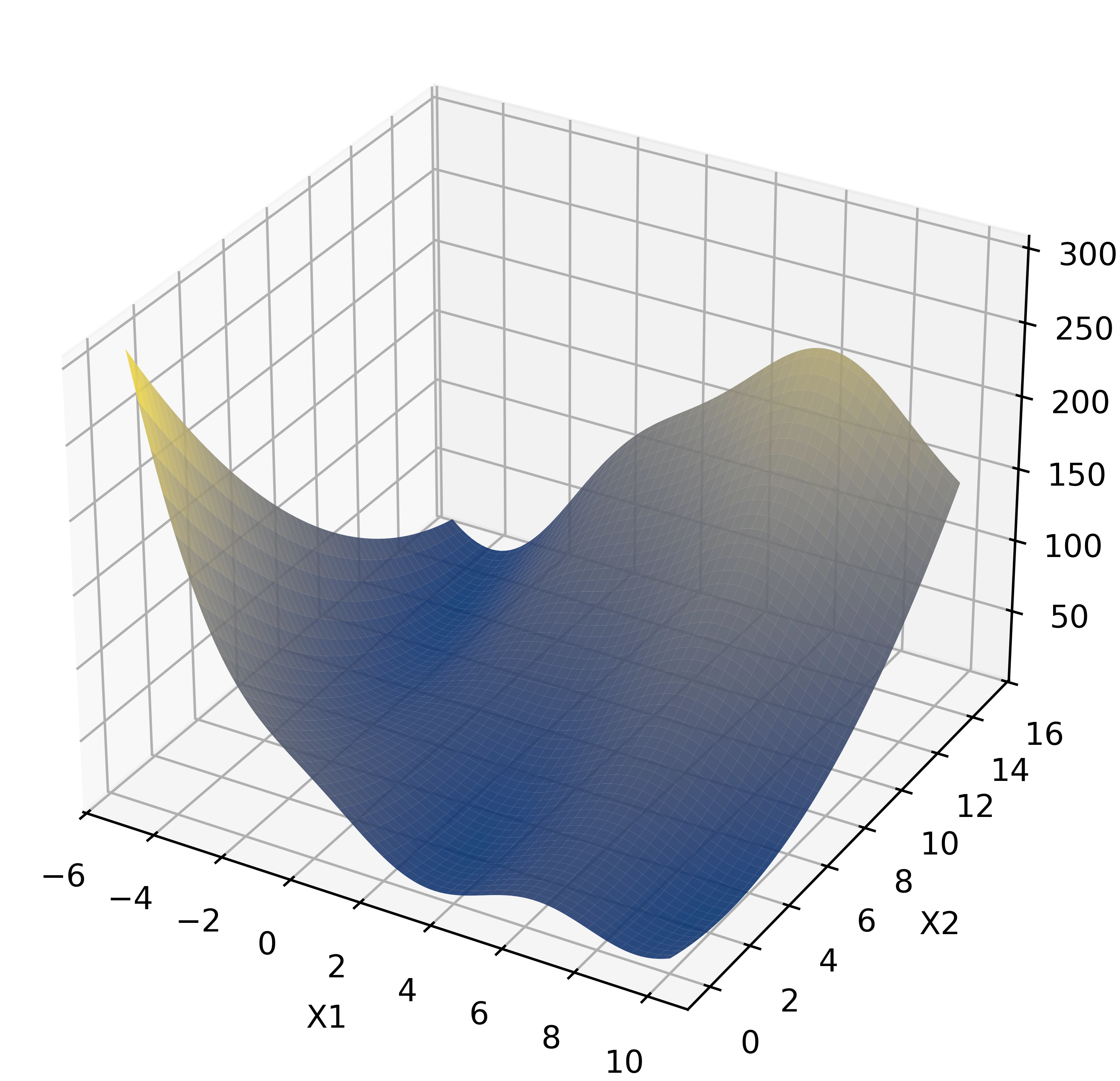}
    \caption{Prediction model($\hat{f}$)}
    \label{fig:function_model}
  \end{subfigure}\hfill
  \begin{subfigure}[t]{0.32\linewidth}
    \centering
    \includegraphics[width=\linewidth]{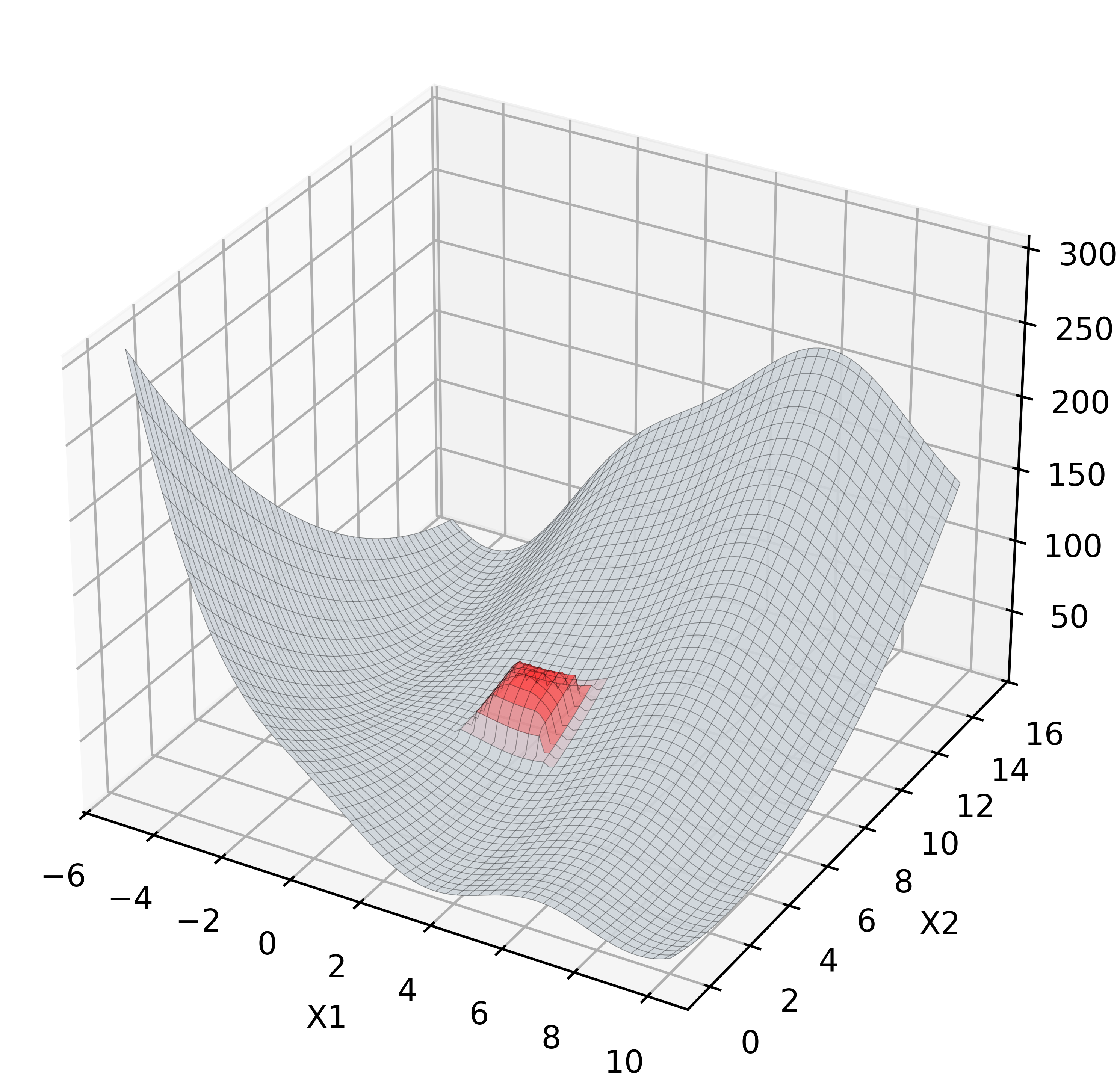}
    \caption{Concept drift($\Delta=2.0$ and $\pdrift=2.0\%$) at $(x_1,x_2)=(2,7)$}
    \label{fig:function_drift}
  \end{subfigure}
  \caption{Experiments setting example of Branin function}
  \label{fig:function_image}
\end{figure}

Figure~\ref{fig:function_image} visualizes the simulation setup on Branin function as an example. Panel~(a) shows the baseline sample $\mathcal{D}_0$ used to fit the surrogate $\hat f$ and to compute the initial residual weights for exploitation. Panel~(b) displays the fitted surface ($\hat f$) over the input domain. Panel~(c) illustrates a localized concept drift, implemented as a constant drift of magnitude $\Delta$ within a small hyper-rectangular region of volume ratio $\pdrift$; this is the ground truth the monitoring approach is meant to detect and localize.

\subsection{Experimental Results}
\label{sub:results}

\begin{figure}[!htb]
  \centering
  \begin{subfigure}[t]{0.33\linewidth}
    \centering
    \includegraphics[width=\linewidth]{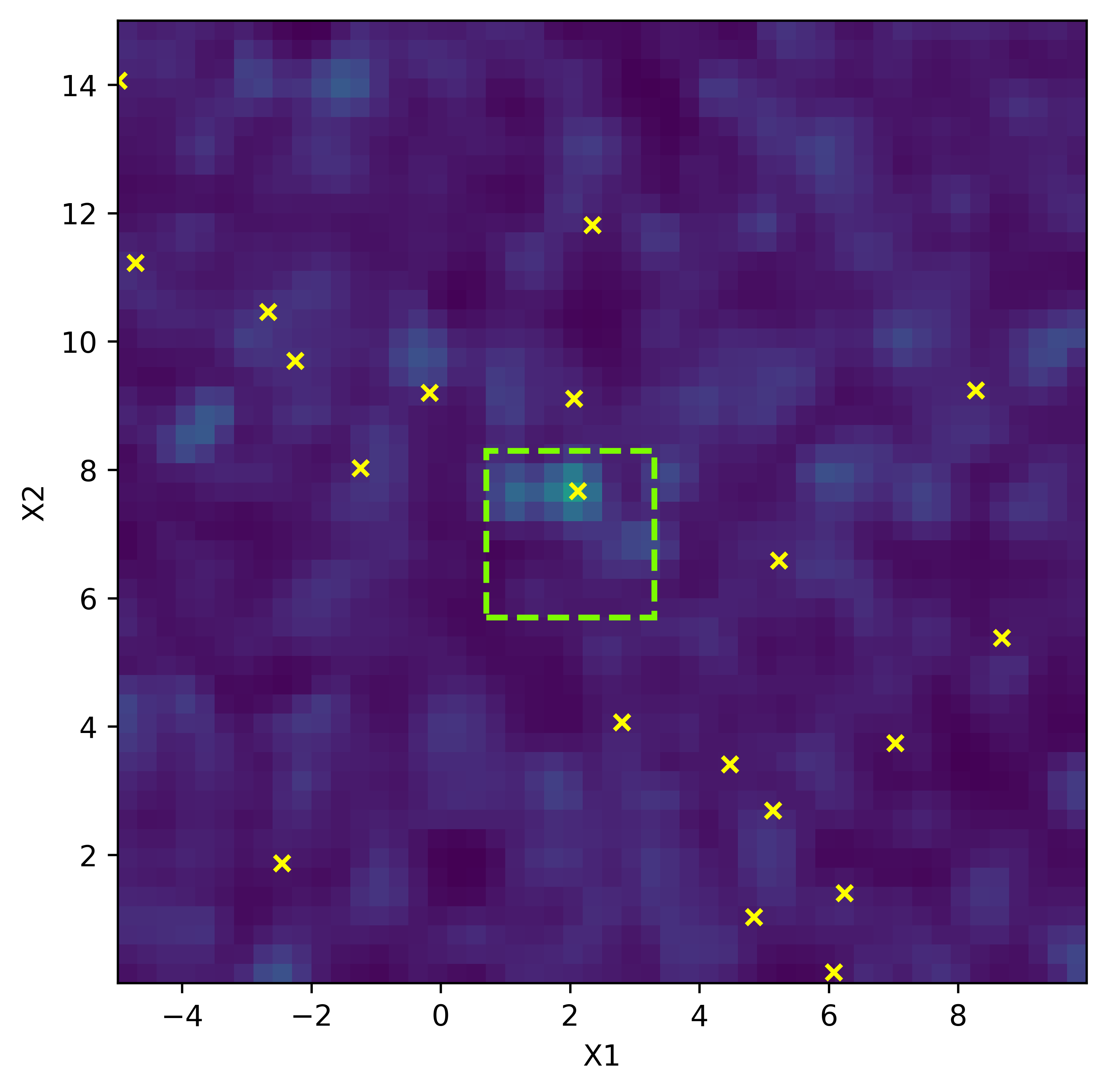}
    \caption{a suspicious hot-spot emerges ($t=33$)}
  \end{subfigure}\hfill
  \begin{subfigure}[t]{0.33\linewidth}
    \centering
    \includegraphics[width=\linewidth]{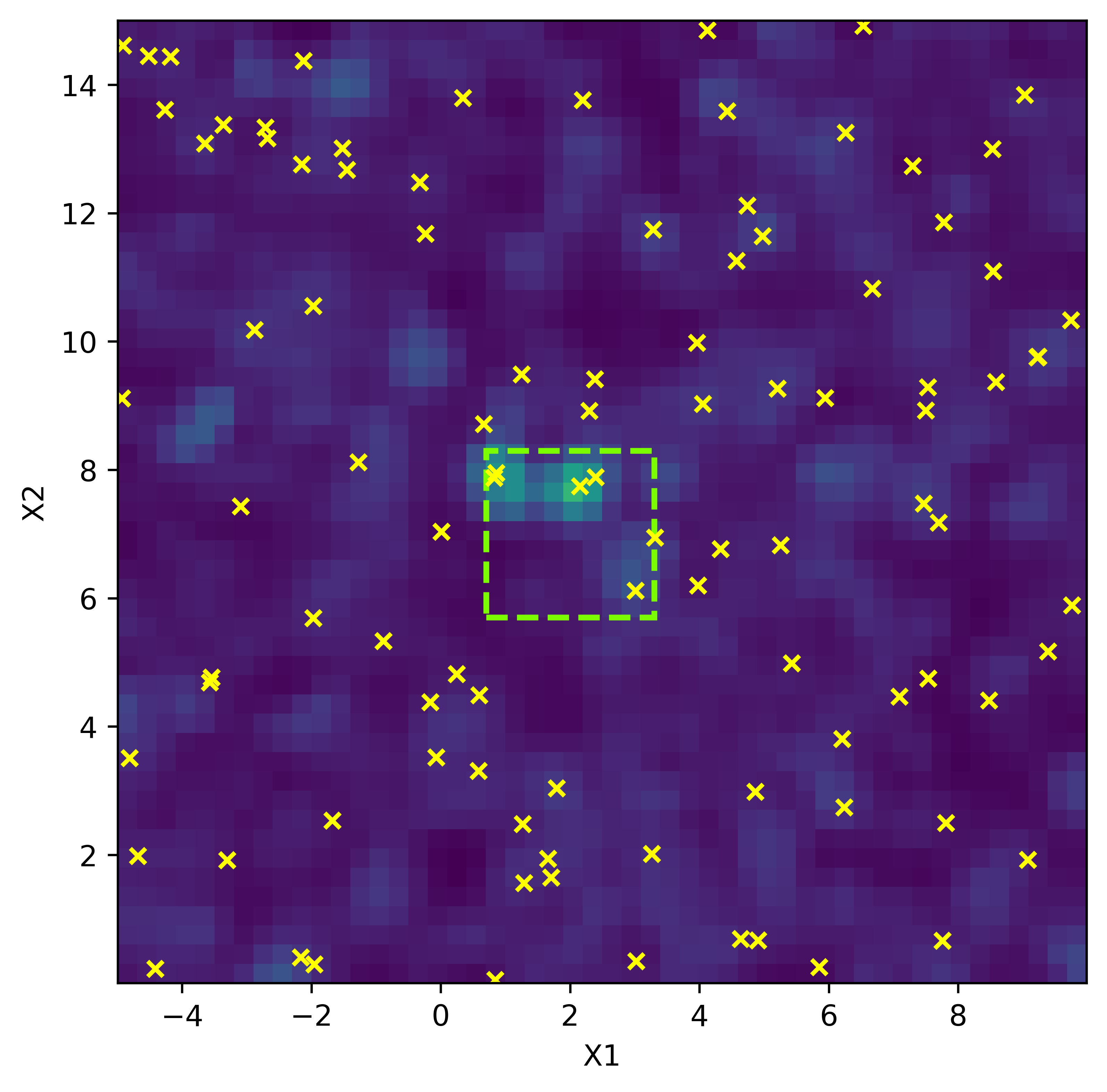}
    \caption{sampling begins to concentrate ($t=34\sim38$)}
  \end{subfigure}\hfill
  \begin{subfigure}[t]{0.33\linewidth}
    \centering
    \includegraphics[width=\linewidth]{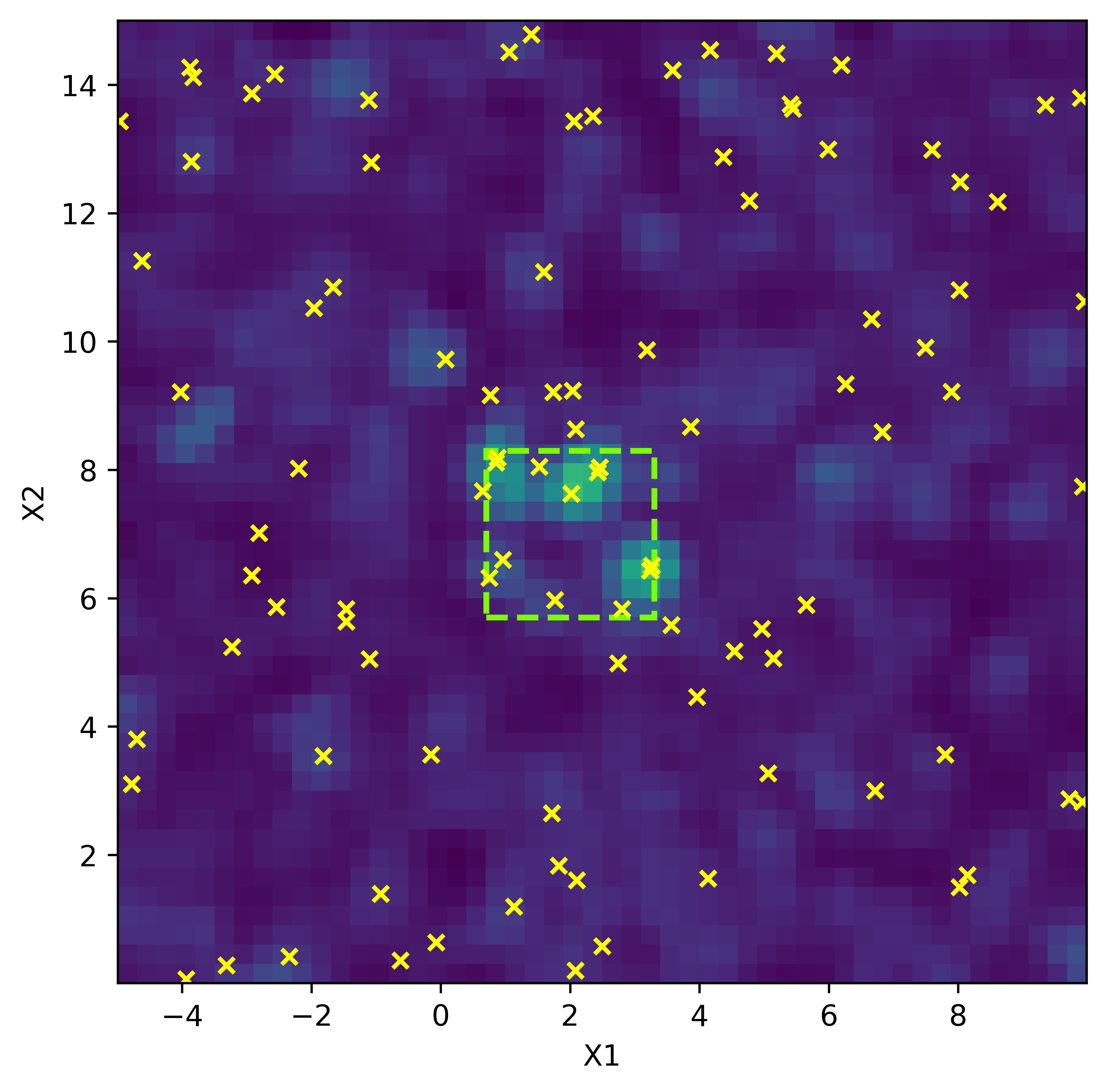}
    \caption{drift region fully localized ($t=39\sim43$)}
  \end{subfigure}
  \caption{Spatial evolution of residual-based sampling around an abrupt drift (true change at $t=30$). Heatmap brightness indicates residual weight; yellow crosses are queried points; the green dashed box marks the true drift region. Snapshots at $t=33,34\sim38$, and $39\sim43$.}
  \label{fig:sampling_evolution}
\end{figure}

In Figure~\ref{fig:sampling_evolution}, consistent with the design in Section~\ref{sec:meth}, exploitation (inverse transform sampling near high residual anchors) and exploration (accept--reject revisits of long-unvisited cells) jointly steer the budget. Before the change, the weight field is diffuse. Soon after the change a localized hot-spot appears, and by $t{=}38$ queries concentrate around it; around this stage the one-sided EWMAs typically cross their UCLs and signal. Meanwhile, any budget not drawn into the hot-spot is allocated by exploration to revisit stale cells and to probe secondary suspects, preserving coverage. Letting the process run a little longer (e.g., to $t{=}43$) further tightens sampling around the affected area, making the drift location clear for diagnosis and model update.

We also compared our proposed framework with two baseline methods:
\begin{itemize}
    \item Random sampling: allocates the sampling budget uniformly throughout the entire input domain, without considering the sampling history.
    \item Score vector ~\citep{zhang2023concept}: monitors the deviation of model parameters using Hotelling's $T^2$ statistic based on score vectors—gradients of the log-likelihood function. Although their proposed method is not adaptive, for fair comparison, we additionally implemented an adaptive sampling variant of the score vector method by incorporating an exploration-exploitation trade-off using $ \epsilon = 0.5 $.
\end{itemize}

Across all conditions, we calibrated control limits to achieve a one-sided in-control
ARL$_0{=}200$. For \ac{acronym} and Random sampling, monitoring used the two one-sided EWMAs introduced in Subsection~\ref{subsec:monitoring} (the top-$r$ absolute-residual mean and the log-variance chart). For the Score vector benchmark, we follow the original paper and use a multivariate exponentially weighted moving average (MEWMA) chart on the score vector with Hotelling-type scaling, calibrating its UCL to ARL$_0{=}200$ \citep{zhang2023concept}. All downstream out-of-control average run length (ARL$_1$) results reported in Subsection~\ref{subsec:abrupt} and Subsection~\ref{subsec:incremental} were obtained under these calibrated limits. Unless otherwise noted, each $(\pdrift,\Delta)$ setting is evaluated over 100 Monte Carlo replications; ARL$_1$ is summarized by the mean with 95\% confidence intervals (CIs) computed as $\pm1.96$ times the standard error (SE) across replications.

\subsubsection{Abrupt Concept Drift}
\label{subsec:abrupt}

The first set of experiments simulated \emph{abrupt} concept drift, where a sudden change in the conditional distribution $P(Y\mid X)$ occurs at a specific time point, set to $t=30$ after initial model training. Here we fix the drift ratio at $\pdrift=1.0\%$ and vary the drift magnitude $\Delta$ from $1.0$ to $3.0$ times the inherent noise level $\sigma_{\text{noise}}$. Detection performance is reported using ARL$_1$, which indicates the average number of observations required to signal drift after its occurrence.

\begin{figure}[t]
\centering
\includegraphics[width=\textwidth]{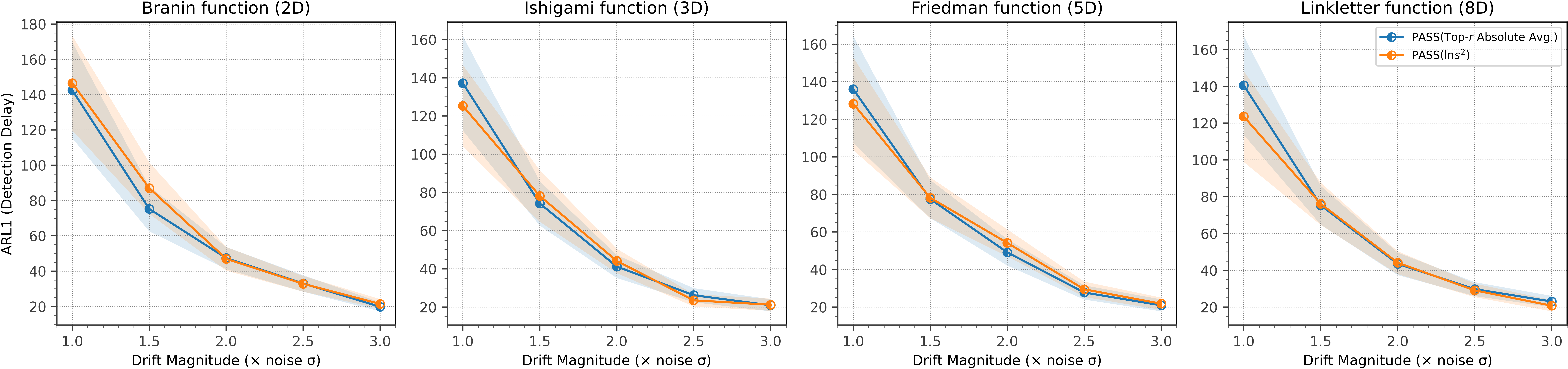}
\caption{Localized abrupt drift with PASS ($\pdrift=1\%$): ARL$_1$ versus drift magnitude $\Delta$ for the top-$r$ absolute-residual and log-variance EWMAs. The two curves track closely.}
\label{fig:avg_vs_var}
\end{figure}

We first compare the two EWMA monitors when coupled with the PASS sampling policy. Figure~\ref{fig:avg_vs_var} displays ARL$_1$ as a function of the drift magnitude $\Delta$ under a fixed drift ratio $\pi_d=1\%$. With ARL$_0$ matched across monitoring methods, the top-$r$ absolute-residual average $A_t^{(r)}$ and the log-variance statistic $V_t=\ln s_t^2$ produce strikingly similar response curves, where ARL$_1$ declines at essentially the same rate as $\Delta$ increases. At $\Delta=1.0$, the variance-based chart yields a slightly smaller mean ARL$_1$, yet the two confidence bands overlap, indicating no statistically meaningful gap at the 95\% level. These findings emphasize two points. First, as motivated in the methodology, both $A_t^{(r)}$ and $V_t$ are effective for detecting localized concept drift. The former concentrates on the largest residuals, while the latter captures dispersion inflation, and each reacts promptly as the severity of drift grows. Second, the adaptive sampling in PASS is robust to the choice of the monitoring statistics, provided they are sufficiently sensitive to localized changes.

\begin{figure}[t]
\centering
\includegraphics[width=\textwidth]{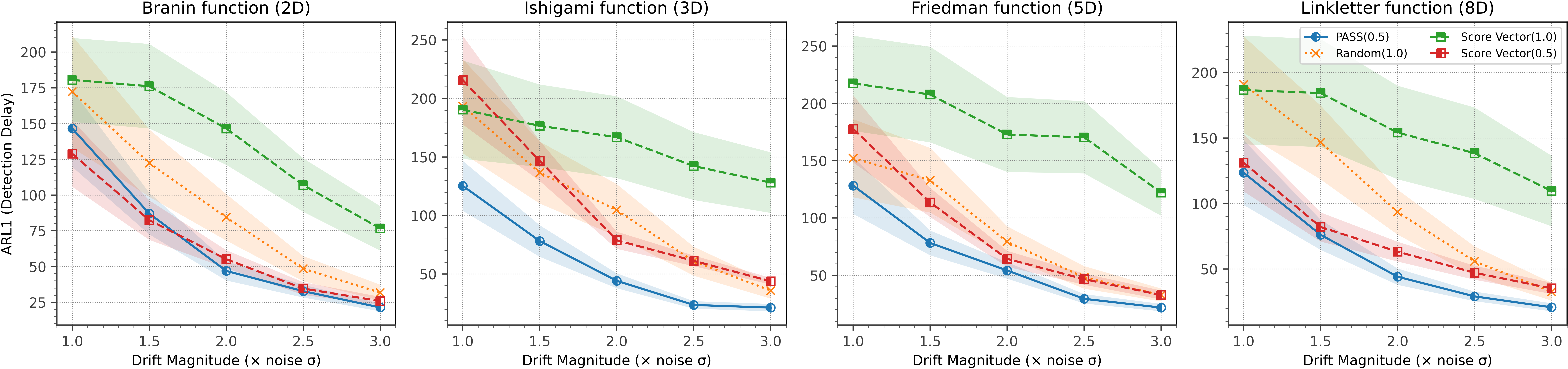}
\caption{Abrupt drift at drift ratio $\pdrift=1.0\%$: ARL$_1$ versus drift magnitude $\Delta$ (in units of $\sigma_{\text{noise}}$) for \ac{acronym} ($\epsilon=0.5$, $V_t$ EWMA), Score vector ($\epsilon=1.0,0.5$), and Random ($V_t$ EWMA). Shaded regions denote 95\% CIs.}
\label{fig:abrupt_strategies}
\end{figure}

Figure~\ref{fig:abrupt_strategies} depicts ARL$_1$ as a function of the drift magnitude $\Delta$ at a fixed drift ratio $\pdrift=1.0\%$ for four representative settings: PASS ($\epsilon=0.5$; $V_t$ EWMA), Score vector with $\epsilon\in\{1.0,0.5\}$, and Random sampling ($\epsilon=1.0$; $V_t$ EWMA). As can be seen from the figure, our proposed PASS delivers the shortest detection delays across settings and functions. Relative to Random sampling, PASS reduces ARL$_1$ by \textbf{38.62\%} on average. The Score vector method with $\epsilon=1.0$ trails even Random sampling in many cases and, more importantly, its ARL$_1$ curve declines only mildly with $\Delta$. This behavior is expected under \emph{localized} drift, where small-area changes that does not correspond to a simple parameter shift of the global model, so the score monitoring statistic can be insensitive, producing a flatter ARL$_1$–$\Delta$ profile. In contrast, the proposed approach remains effective independent of model dimensionality, rapidly lowering ARL$_1$ as $\Delta$ increases.

We also evaluated a hybrid that embeds our adaptive sampling policy into the Score vector monitor. This hybrid yields a \textbf{50.38\%} average reduction in ARL$_1$ relative to the plain Score vector with $\epsilon=1.0$. On the Branin function, for $\Delta<2.0$ the hybrid performs nearly on par with our $V_t$ EWMA. Together with the average–variance comparison, these results indicate that the gains are driven mainly by the adaptive sampling: the framework is robust to the specific choice of monitoring chart while remaining highly effective for localized concept drift.

\begin{table}[!htb]
\centering
\renewcommand{\arraystretch}{1.2}
\caption{Sensitivity analysis in the case of abrupt drift under the $V_t$ EWMA monitor (log–variance): ARL$_1$ mean(SE) across drift ratios $\pdrift$, drift magnitudes $\Delta$, and exploration rate $\epsilon$ for Branin (2D), Ishigami (3D), Friedman (5D), and Linkletter (8D).}
\label{tab:abrupt_sensitivity}
\resizebox{\textwidth}{!}{
\begin{tabular}{cc|rrr|rrr}
\hline
\multirow{2}{*}{$\pdrift$} & \multirow{2}{*}{$\Delta(\times\sigma)$} &
\multicolumn{3}{c|}{\textbf{Branin (2D)}} &
\multicolumn{3}{c}{\textbf{Ishigami (3D)}} \\
& & $\epsilon=0.2$ & $\epsilon=0.5$ & $\epsilon=0.8$ & $\epsilon=0.2$ & $\epsilon=0.5$ & $\epsilon=0.8$ \\
\hline
\multirow{4}{*}{1.0\%} & 1.0  & \best{\meanSE{120.15}{9.63}} & \meanSE{146.57}{13.70} & \meanSE{128.23}{11.76} & \best{\meanSE{115.45}{8.93}} & \meanSE{125.34}{10.91} & \meanSE{149.52}{13.75} \\
 & 1.5  & \best{\meanSE{67.74}{4.13}} & \meanSE{87.04}{7.44} & \meanSE{93.41}{8.93} & \best{\meanSE{70.94}{4.79}} & \meanSE{78.17}{6.88} & \meanSE{124.41}{11.52} \\
 & 2.0  & \best{\meanSE{40.86}{2.40}} & \meanSE{46.95}{3.42} & \meanSE{63.96}{5.79} & \best{\meanSE{39.54}{2.50}} & \meanSE{44.17}{3.25} & \meanSE{67.62}{5.65} \\
 & 2.5  & \best{\meanSE{25.24}{1.64}} & \meanSE{32.78}{2.28} & \meanSE{39.48}{3.18} & \meanSE{26.30}{1.70} & \best{\meanSE{23.53}{1.64}} & \meanSE{43.09}{3.18} \\
 & 3.0  & \best{\meanSE{17.78}{1.18}} & \meanSE{21.39}{1.38} & \meanSE{24.98}{1.89} & \best{\meanSE{15.50}{1.07}} & \meanSE{21.24}{1.61} & \meanSE{28.11}{2.19} \\
\hline
\multirow{4}{*}{2.0\%} & 1.0  & \best{\meanSE{93.67}{7.52}} & \meanSE{128.90}{9.95} & \meanSE{122.93}{12.02} & \best{\meanSE{88.38}{7.08}} & \meanSE{89.33}{8.37} & \meanSE{133.88}{10.68} \\
 & 1.5  & \best{\meanSE{43.12}{2.82}} & \meanSE{58.29}{4.70} & \meanSE{69.38}{8.36} & \best{\meanSE{50.67}{3.24}} & \meanSE{53.17}{4.49} & \meanSE{87.04}{7.91} \\
 & 2.0  & \best{\meanSE{26.25}{1.63}} & \meanSE{28.35}{1.97} & \meanSE{39.61}{3.24} & \best{\meanSE{25.16}{1.69}} & \meanSE{28.09}{2.26} & \meanSE{38.86}{3.17} \\
 & 2.5  & \best{\meanSE{14.92}{1.01}} & \meanSE{17.54}{1.31} & \meanSE{20.83}{1.80} & \best{\meanSE{14.71}{1.16}} & \meanSE{15.87}{1.03} & \meanSE{23.79}{1.74} \\
 & 3.0  & \best{\meanSE{10.22}{0.77}} & \meanSE{10.87}{0.82} & \meanSE{13.90}{1.27} & \best{\meanSE{9.48}{0.62}} & \meanSE{10.42}{0.69} & \meanSE{13.52}{1.33} \\
\hline
\multirow{4}{*}{3.0\%} & 1.0  & \best{\meanSE{76.41}{5.20}} & \meanSE{89.47}{8.15} & \meanSE{122.81}{9.75} & \best{\meanSE{71.10}{6.12}} & \meanSE{77.90}{7.36} & \meanSE{107.82}{15.70} \\
 & 1.5  & \best{\meanSE{36.09}{2.40}} & \meanSE{40.26}{3.50} & \meanSE{49.93}{4.08} & \best{\meanSE{36.53}{2.46}} & \meanSE{37.58}{2.80} & \meanSE{48.25}{3.81} \\
 & 2.0  & \best{\meanSE{19.17}{1.30}} & \meanSE{20.81}{1.63} & \meanSE{25.49}{2.09} & \meanSE{19.53}{1.27} & \best{\meanSE{18.88}{1.69}} & \meanSE{29.56}{2.46} \\
 & 2.5  & \best{\meanSE{10.60}{0.76}} & \meanSE{10.76}{0.82} & \meanSE{14.61}{1.33} & \best{\meanSE{9.89}{0.70}} & \meanSE{12.73}{1.03} & \meanSE{15.12}{1.15} \\
 & 3.0  & \best{\meanSE{7.01}{0.52}} & \meanSE{7.55}{0.54} & \meanSE{8.98}{0.84} & \best{\meanSE{6.96}{0.49}} & \meanSE{7.07}{0.56} & \meanSE{9.60}{0.92} \\
\hline

\end{tabular}
}

\vspace{4pt}

\resizebox{\textwidth}{!}{
\begin{tabular}{cc|rrr|rrr}
\hline
\multirow{2}{*}{$\pdrift$} & \multirow{2}{*}{$\Delta(\times\sigma)$} &
\multicolumn{3}{c|}{\textbf{Friedman (5D)}} &
\multicolumn{3}{c}{\textbf{Linkletter (8D)}} \\
& & $\epsilon=0.2$ & $\epsilon=0.5$ & $\epsilon=0.8$ & $\epsilon=0.2$ & $\epsilon=0.5$ & $\epsilon=0.8$ \\
\hline
\multirow{4}{*}{1.0\%} & 1.0  & \best{\meanSE{105.31}{8.51}} & \meanSE{128.28}{12.61} & \meanSE{167.97}{14.94} & \best{\meanSE{110.44}{8.71}} & \meanSE{123.60}{12.62} & \meanSE{180.12}{18.65} \\
 & 1.5  & \best{\meanSE{63.51}{4.61}} & \meanSE{78.21}{5.53} & \meanSE{108.08}{8.69} & \best{\meanSE{67.57}{4.45}} & \meanSE{76.14}{5.88} & \meanSE{129.06}{12.28} \\
 & 2.0  & \best{\meanSE{38.59}{2.87}} & \meanSE{54.21}{3.67} & \meanSE{64.79}{4.48} & \best{\meanSE{43.36}{2.85}} & \meanSE{44.11}{3.18} & \meanSE{71.25}{5.88} \\
 & 2.5  & \best{\meanSE{27.63}{2.00}} & \meanSE{29.47}{2.09} & \meanSE{43.60}{2.54} & \best{\meanSE{24.84}{1.71}} & \meanSE{29.10}{1.88} & \meanSE{45.57}{2.97} \\
 & 3.0  & \best{\meanSE{17.93}{1.50}} & \meanSE{21.80}{1.66} & \meanSE{28.83}{2.11} & \best{\meanSE{19.22}{1.47}} & \meanSE{20.75}{1.56} & \meanSE{29.41}{2.15} \\
\hline
\multirow{4}{*}{2.0\%} & 1.0  & \best{\meanSE{96.87}{7.40}} & \meanSE{104.25}{8.93} & \meanSE{138.99}{12.34} & \best{\meanSE{88.51}{6.68}} & \meanSE{97.59}{9.02} & \meanSE{123.63}{12.28} \\
 & 1.5  & \best{\meanSE{44.74}{3.32}} & \meanSE{55.30}{3.51} & \meanSE{75.28}{5.67} & \meanSE{52.07}{3.45} & \best{\meanSE{51.31}{3.75}} & \meanSE{81.48}{7.83} \\
 & 2.0  & \best{\meanSE{23.20}{1.74}} & \meanSE{28.74}{2.04} & \meanSE{40.89}{2.95} & \best{\meanSE{28.13}{2.11}} & \meanSE{30.32}{2.17} & \meanSE{36.87}{2.89} \\
 & 2.5  & \best{\meanSE{16.08}{1.23}} & \meanSE{19.02}{1.35} & \meanSE{26.90}{1.82} & \meanSE{17.13}{1.39} & \best{\meanSE{15.88}{1.06}} & \meanSE{23.72}{1.77} \\
 & 3.0  & \best{\meanSE{10.46}{0.96}} & \meanSE{11.20}{0.94} & \meanSE{15.01}{1.10} & \meanSE{10.96}{0.96} & \best{\meanSE{10.65}{0.86}} & \meanSE{13.73}{1.15} \\
\hline
\multirow{4}{*}{3.0\%} & 1.0  & \best{\meanSE{69.87}{5.87}} & \meanSE{89.99}{7.60} & \meanSE{93.55}{8.32} & \best{\meanSE{81.50}{5.43}} & \meanSE{82.86}{7.51} & \meanSE{113.75}{12.02} \\
 & 1.5  & \best{\meanSE{35.47}{2.58}} & \meanSE{42.46}{3.00} & \meanSE{55.59}{4.23} & \best{\meanSE{36.64}{2.57}} & \meanSE{38.86}{3.12} & \meanSE{49.81}{3.89} \\
 & 2.0  & \best{\meanSE{17.19}{1.43}} & \meanSE{20.72}{1.63} & \meanSE{29.88}{2.20} & \best{\meanSE{21.34}{1.59}} & \meanSE{22.03}{1.67} & \meanSE{25.14}{1.97} \\
 & 2.5  & \best{\meanSE{10.80}{0.95}} & \meanSE{11.92}{0.84} & \meanSE{16.07}{1.25} & \best{\meanSE{11.28}{0.96}} & \meanSE{11.86}{0.91} & \meanSE{15.04}{1.30} \\
 & 3.0  & \best{\meanSE{7.04}{0.55}} & \meanSE{7.24}{0.61} & \meanSE{8.99}{0.63} & \meanSE{6.99}{0.56} & \best{\meanSE{6.79}{0.52}} & \meanSE{8.60}{0.74} \\
\hline
\end{tabular}
}
\end{table}

Complementing Figure~\ref{fig:abrupt_strategies}, Table~\ref{tab:abrupt_sensitivity} summarizes the $\epsilon$–sensitivity of PASS under the $V_t$ EWMA (log–variance) monitoring statistic. Each $(\pdrift,\Delta)$ cell is averaged over 100 replications; entries report the ARL$_1$ mean with SE in parentheses, and the per-row best $\epsilon$ is underlined. Two patterns stand out. First, $\epsilon\in\{0.2,0.5\}$ generally yields smaller ARL$_1$ than $\epsilon=0.8$. The advantage of a smaller $\epsilon$ is most visible when drift is highly localized (small $\pdrift$) or weak ($\Delta$ near 1–1.5): concentrating more of the budget on exploitation secures denser sampling in suspicious areas. Second, as the affected region grows or the magnitude increases, the gaps between $\epsilon$’s narrow. This implies that exploration more frequently lands inside the drift, reducing the benefit of additional exploitation. In short, smaller $\epsilon$ tends to help when drift is highly localized or weak, whereas broader drifts can be detected even with a larger $\epsilon$; if information from other regions is also desired (e.g., when multiple drifting regions are plausible), leaning larger may further aid coverage.


\subsubsection{Incremental Concept Drift}
\label{subsec:incremental}

We also investigated the detection performance under \emph{incremental} concept drift, where the change builds up between $t=30$ and $t=60$. To compare our strategy with the benchmarks, we vary the drift magnitude $\Delta\in\{1.0,1.5,2.0,2.5,3.0\}$ in units of the noise standard deviation $\sigma$ with the affected area ratio of 1.0\%. As in the abrupt setting, the top-$r$ absolute-residual EWMA $A_t^{(r)}$ and the log–variance EWMA $V_t=\ln s_t^2$ produce very similar ARL$_1$ curves with overlapping 95\% intervals; accordingly, the strategy comparison below is reported under the $V_t$ monitor.

\begin{figure}[t]
\centering
\includegraphics[width=\textwidth]{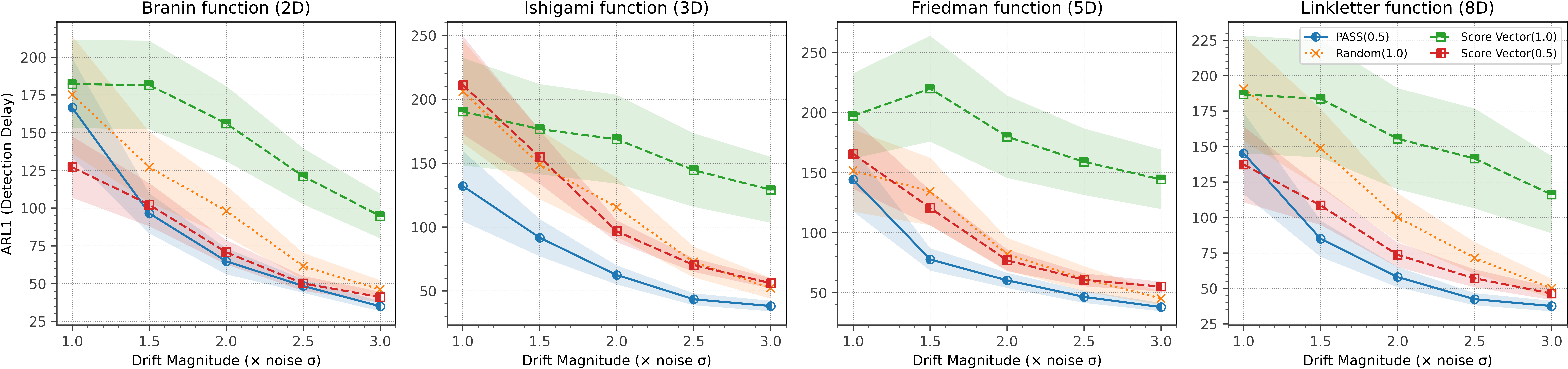}
\caption{Incremental drift at drift ratio $\pdrift=1.0\%$: ARL$_1$ versus drift magnitude $\Delta$ (in units of $\sigma_{\text{noise}}$) for \ac{acronym} ($\epsilon=0.5$, $V_t$ EWMA), Score vector ($\epsilon=1.0,0.5$), and Random ($V_t$ EWMA). Shaded regions denote 95\% CIs.}
\label{fig:strategy_incremental}
\end{figure}

\begin{figure}[t]
\centering
\includegraphics[width=\textwidth]{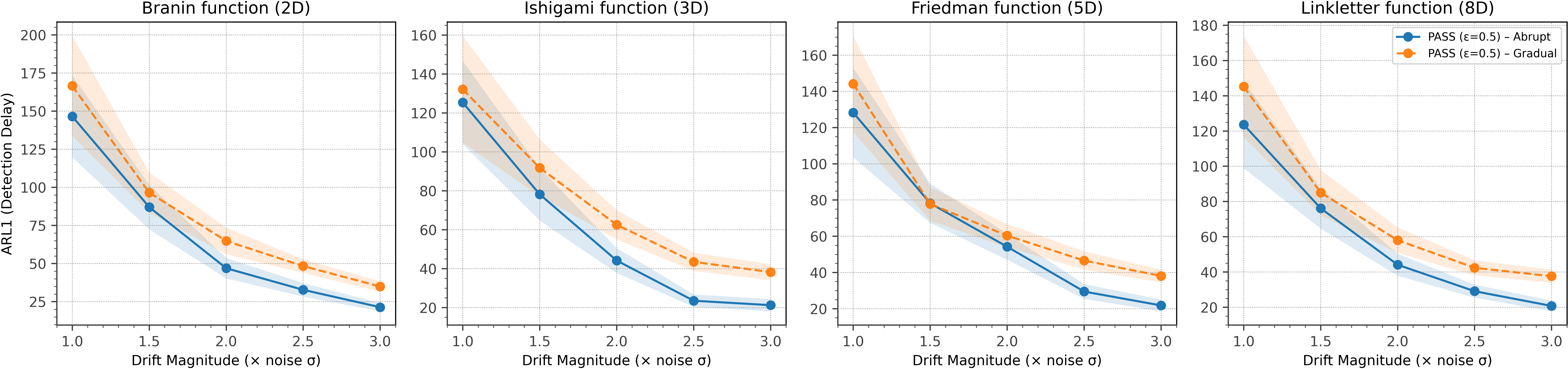}
\caption{PASS ($\epsilon=0.5$, $V_t$ EWMA): ARL$_1$ versus drift magnitude $\Delta$ under abrupt (solid) and incremental (dashed) drift at $\pdrift=1.0\%$. Shaded regions denote 95\% CIs.}
\label{fig:abrupt_vs_gradual}
\end{figure}

As can be seen in Figure~\ref{fig:strategy_incremental}, ARL$_1$ decreases monotonically with $\Delta$ and the shaded 95\% bands tighten as the signal strengthens across functions. The ranking mirrors the abrupt case: PASS ($\epsilon=0.5$) attains the shortest delays, Score vector ($\epsilon=0.5$) generally improves over Random in most panels, and Score($\epsilon=1.0$) lags. Incremental drift is uniformly harder than abrupt one for the same $(\pdrift,\Delta)$, meaning detections are slower because the evidence accrues gradually. This contrast is visible in Figure~\ref{fig:abrupt_vs_gradual}, which overlays PASS ($\epsilon=0.5$, $V_t$) under abrupt (solid) versus incremental (dashed) drift at $\pdrift=1.0\%$; the incremental curve sits consistently above the abrupt curve at each $\Delta$.

\begin{table}[!htb]
\centering
\renewcommand{\arraystretch}{1.2}
\caption{Sensitivity analysis with respect to drift ratios ($\pdrift$), drift magnitudes ($\Delta$), and exploration--exploitation trade-off ($\epsilon$) with incremental drift: ARL$_1$ mean(SE) for Branin (2D), Ishigami (3D), Friedman (5D), and Linkletter (8D).}
\label{tab:incremental_sensitivity}
\resizebox{\textwidth}{!}{
\begin{tabular}{cc|rrr|rrr}
\hline
\multirow{2}{*}{$\pdrift$} & \multirow{2}{*}{$\Delta(\times\sigma)$} &
\multicolumn{3}{c|}{\textbf{Branin (2D)}} &
\multicolumn{3}{c}{\textbf{Ishigami (3D)}} \\
& & $\epsilon=0.2$ & $\epsilon=0.5$ & $\epsilon=0.8$ & $\epsilon=0.2$ & $\epsilon=0.5$ & $\epsilon=0.8$ \\
\hline
\multirow{4}{*}{1.0\%} & 1.0  & \best{\meanSE{132.77}{10.17}} & \meanSE{166.55}{16.80} & \meanSE{145.31}{13.87} & \meanSE{142.30}{13.48} & \best{\meanSE{132.13}{14.04}} & \meanSE{203.61}{17.37} \\
 & 1.5  & \best{\meanSE{79.20}{5.00}} & \meanSE{96.56}{6.75} & \meanSE{123.09}{10.44} & \best{\meanSE{84.14}{4.88}} & \meanSE{91.75}{7.43} & \meanSE{118.19}{10.78} \\
 & 2.0  & \best{\meanSE{56.33}{2.86}} & \meanSE{64.85}{4.42} & \meanSE{78.76}{5.68} & \best{\meanSE{51.78}{2.68}} & \meanSE{62.55}{3.84} & \meanSE{81.65}{5.39} \\
 & 2.5  & \best{\meanSE{42.46}{1.89}} & \meanSE{48.38}{2.32} & \meanSE{62.56}{3.81} & \best{\meanSE{43.20}{1.91}} & \meanSE{43.49}{2.37} & \meanSE{59.26}{3.04} \\
 & 3.0  & \best{\meanSE{34.35}{1.34}} & \meanSE{34.95}{1.69} & \meanSE{40.81}{2.27} & \best{\meanSE{35.84}{1.48}} & \meanSE{38.23}{2.10} & \meanSE{44.58}{2.25} \\
\hline
\multirow{4}{*}{2.0\%} & 1.0  & \best{\meanSE{94.90}{7.60}} & \meanSE{130.54}{11.04} & \meanSE{126.66}{11.06} & \best{\meanSE{91.47}{6.69}} & \meanSE{102.31}{9.87} & \meanSE{154.91}{13.02} \\
 & 1.5  & \best{\meanSE{59.52}{3.37}} & \meanSE{68.04}{4.56} & \meanSE{81.01}{5.98} & \best{\meanSE{61.07}{3.41}} & \meanSE{67.90}{4.10} & \meanSE{83.27}{5.88} \\
 & 2.0  & \best{\meanSE{40.36}{1.72}} & \meanSE{46.33}{2.30} & \meanSE{56.89}{3.26} & \best{\meanSE{42.36}{2.05}} & \meanSE{45.81}{2.48} & \meanSE{62.14}{3.29} \\
 & 2.5  & \best{\meanSE{31.67}{1.16}} & \meanSE{35.81}{1.57} & \meanSE{38.72}{1.84} & \best{\meanSE{31.08}{1.32}} & \meanSE{32.06}{1.49} & \meanSE{42.54}{2.12} \\
 & 3.0  & \best{\meanSE{27.51}{0.95}} & \meanSE{28.41}{1.05} & \meanSE{32.38}{1.43} & \meanSE{27.67}{0.95} & \best{\meanSE{26.45}{1.02}} & \meanSE{35.36}{1.38} \\
\hline
\multirow{4}{*}{3.0\%} & 1.0  & \best{\meanSE{94.63}{6.24}} & \meanSE{107.38}{9.53} & \meanSE{111.96}{8.94} & \meanSE{96.88}{6.82} & \best{\meanSE{96.15}{8.74}} & \meanSE{132.11}{11.54} \\
 & 1.5  & \best{\meanSE{47.96}{2.32}} & \meanSE{52.30}{2.91} & \meanSE{65.91}{4.47} & \meanSE{49.94}{2.46} & \best{\meanSE{46.83}{2.80}} & \meanSE{64.00}{4.36} \\
 & 2.0  & \meanSE{38.43}{1.61} & \best{\meanSE{37.77}{1.60}} & \meanSE{39.46}{2.08} & \meanSE{37.02}{1.76} & \best{\meanSE{36.88}{1.88}} & \meanSE{42.46}{1.94} \\
 & 2.5  & \best{\meanSE{29.09}{1.06}} & \meanSE{29.74}{1.28} & \meanSE{31.78}{1.38} & \best{\meanSE{26.22}{0.97}} & \meanSE{29.77}{1.22} & \meanSE{33.90}{1.46} \\
 & 3.0  & \meanSE{24.93}{0.71} & \best{\meanSE{24.34}{0.84}} & \meanSE{25.68}{1.11} & \meanSE{25.19}{0.82} & \best{\meanSE{23.47}{0.91}} & \meanSE{29.31}{0.96} \\
\hline
\end{tabular}
}

\vspace{4pt}

\resizebox{\textwidth}{!}{
\begin{tabular}{cc|rrr|rrr}
\hline
\multirow{2}{*}{$\pdrift$} & \multirow{2}{*}{$\Delta(\times\sigma)$} &
\multicolumn{3}{c|}{\textbf{Friedman (5D)}} &
\multicolumn{3}{c}{\textbf{Linkletter (8D)}} \\
& & $\epsilon=0.2$ & $\epsilon=0.5$ & $\epsilon=0.8$ & $\epsilon=0.2$ & $\epsilon=0.5$ & $\epsilon=0.8$ \\
\hline
\multirow{4}{*}{1.0\%} & 1.0  & \best{\meanSE{117.66}{8.32}} & \meanSE{144.18}{13.55} & \meanSE{166.64}{14.77} & \best{\meanSE{115.71}{8.66}} & \meanSE{145.28}{14.69} & \meanSE{171.21}{16.27} \\
 & 1.5  & \best{\meanSE{76.37}{5.24}} & \meanSE{77.83}{4.76} & \meanSE{102.15}{7.93} & \best{\meanSE{74.01}{4.27}} & \meanSE{85.07}{6.53} & \meanSE{124.28}{11.28} \\
 & 2.0  & \best{\meanSE{53.87}{3.12}} & \meanSE{60.35}{3.27} & \meanSE{78.82}{4.96} & \best{\meanSE{57.78}{3.13}} & \meanSE{58.04}{3.70} & \meanSE{86.66}{6.82} \\
 & 2.5  & \best{\meanSE{43.21}{2.49}} & \meanSE{46.58}{2.57} & \meanSE{57.90}{3.34} & \best{\meanSE{42.22}{1.93}} & \meanSE{42.40}{2.17} & \meanSE{60.33}{3.39} \\
 & 3.0  & \best{\meanSE{33.46}{1.62}} & \meanSE{38.09}{1.78} & \meanSE{47.11}{2.12} & \best{\meanSE{35.76}{1.39}} & \meanSE{37.60}{1.89} & \meanSE{48.96}{2.57} \\
\hline
\multirow{4}{*}{2.0\%} & 1.0  & \best{\meanSE{97.86}{6.72}} & \meanSE{108.79}{8.56} & \meanSE{128.46}{11.10} & \best{\meanSE{101.33}{7.82}} & \meanSE{104.51}{9.75} & \meanSE{150.38}{13.90} \\
 & 1.5  & \meanSE{65.50}{3.87} & \best{\meanSE{65.23}{3.79}} & \meanSE{77.94}{5.65} & \meanSE{63.45}{3.54} & \best{\meanSE{61.54}{3.93}} & \meanSE{93.02}{8.17} \\
 & 2.0  & \best{\meanSE{41.49}{1.97}} & \meanSE{47.81}{2.45} & \meanSE{53.32}{3.17} & \best{\meanSE{43.47}{2.02}} & \meanSE{44.31}{2.46} & \meanSE{59.27}{3.76} \\
 & 2.5  & \best{\meanSE{32.10}{1.39}} & \meanSE{36.38}{1.57} & \meanSE{40.40}{1.69} & \best{\meanSE{33.73}{1.37}} & \meanSE{33.96}{1.58} & \meanSE{43.73}{2.22} \\
 & 3.0  & \best{\meanSE{27.52}{1.07}} & \meanSE{29.84}{1.03} & \meanSE{34.33}{1.39} & \meanSE{28.83}{1.00} & \best{\meanSE{27.91}{1.14}} & \meanSE{33.38}{1.42} \\
\hline
\multirow{4}{*}{3.0\%} & 1.0  & \best{\meanSE{92.46}{6.50}} & \meanSE{110.61}{8.87} & \meanSE{117.93}{9.40} & \meanSE{86.10}{6.17} & \best{\meanSE{83.30}{6.62}} & \meanSE{129.48}{11.64} \\
 & 1.5  & \best{\meanSE{50.05}{2.75}} & \meanSE{55.16}{2.83} & \meanSE{63.76}{4.23} & \best{\meanSE{52.45}{2.83}} & \meanSE{52.84}{3.35} & \meanSE{67.25}{4.38} \\
 & 2.0  & \best{\meanSE{34.90}{1.56}} & \meanSE{40.10}{1.97} & \meanSE{43.94}{2.07} & \meanSE{37.39}{1.62} & \best{\meanSE{36.17}{1.76}} & \meanSE{42.85}{2.26} \\
 & 2.5  & \best{\meanSE{28.11}{1.06}} & \meanSE{29.64}{1.15} & \meanSE{33.89}{1.31} & \meanSE{28.79}{1.08} & \best{\meanSE{27.74}{1.10}} & \meanSE{33.21}{1.49} \\
 & 3.0  & \best{\meanSE{23.85}{0.87}} & \meanSE{24.68}{0.92} & \meanSE{28.21}{0.95} & \meanSE{25.49}{0.90} & \best{\meanSE{24.18}{0.87}} & \meanSE{27.84}{0.99} \\
\hline
\end{tabular}
}
\end{table}

Table~\ref{tab:incremental_sensitivity} provides $\epsilon$–sensitivity under the $V_t$ monitoring statistic, analogous to the abrupt case. As before, $\epsilon\in\{0.2,0.5\}$ typically attains the smallest ARL$_1$, especially when drift is highly localized (small $\pdrift$) or weak ($\Delta$ near 1–1.5), where denser exploitation accelerates learning in the suspicious region. At $\pdrift=3\%$, however, the advantage often shifts toward $\epsilon=0.5$. Under incremental changes, the early-stage signal is weak and subtle, so an overly small $\epsilon$ may overconcentrate budget on transient residual spikes, delaying coverage of the broader affected area as the drift grows. Therefore, consistent with the abrupt case, a moderate exploration level is effective; empirically, weaker signals in early stage of drifts make a slightly smaller $\epsilon$ generally more effective, while broader drifts are still adequately covered with larger $\epsilon$.

\section{Case Study}
\label{sec:case}

To further assess the performance of our sampling framework, we carried out a case study using data from the UK electricity market, focusing on electricity prices and solar power generation. Renewable sources like wind and solar are playing an increasingly important role in climate change mitigation, with projections estimating that they will provide around 40\% of global electricity by 2030 \citep{IEA2023}. While this transition supports sustainability goals, it also introduces new challenges—most notably, greater volatility and uncertainty in electricity prices \citep{morales2013integrating}. These challenges have been amplified by recent geopolitical events, particularly disruptions to natural gas supplies, which have further destabilized markets and increased the risk of price spikes \citep{fabra2023reforming}. As the market evolves, the statistical properties of electricity data can change in meaningful ways. These shifts can affect key predictor-target relationships, leading to distribution drift or the emergence of confounding factors. This poses a significant threat to the reliability of predictive models, particularly those trained on historical data under the assumption of system stability. Without proper drift detection and adaptation, models risk producing inaccurate or outdated forecasts.

\begin{figure}[!hbt]
    \centering
    \includegraphics[width=\textwidth]{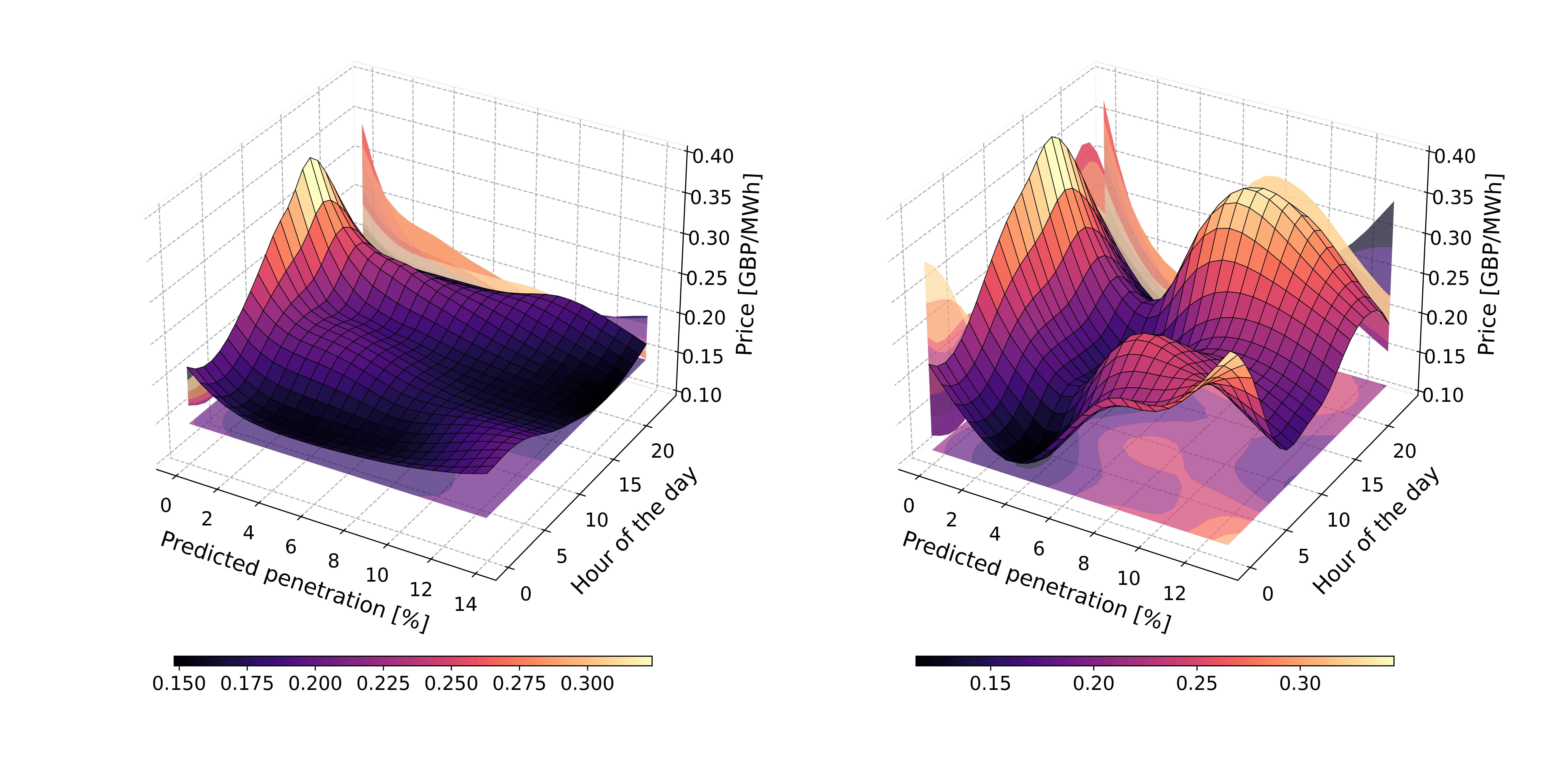}
    \caption{Comparison of electricity price curves between stable market conditions (2020) and energy crisis conditions (2022).}
    \label{fig:lwpr_solar}
\end{figure}

To explore these dynamics, we analyzed electricity market data from two distinct periods: the relatively stable year 2020 and the energy crisis of 2022. Our objective was to understand how structural changes in the market affected the predictive relationship between input features and electricity prices. The dataset included half-hourly observations, with a focus on two key predictors: the hour of the day ($X_1$) and solar power penetration ($X_2$). Hour of day captures daily demand cycles, with price peaks typically occurring in the morning and evening. Solar penetration, defined as the share of generation from solar sources, is critical because it tends to lower prices due to its low marginal cost, displacing more expensive energy sources. The target variable ($Y$) is the day-ahead electricity price from Amsterdam Power Exchange (APX), measured in GBP/MWh.

Under normal market conditions, such as in 2020, we observed a clear inverse relationship between solar power penetration and electricity prices (Figure~\ref{fig:lwpr_solar}, left). This aligns with the merit-order effect, where low-cost renewable energy displaces more expensive fossil fuel generation, leading to lower prices. However, during the 2022 energy crisis, this relationship appeared to weaken or even disappear (Figure~\ref{fig:lwpr_solar}, right). A combination of factors—including geopolitical instability from the war in Ukraine, reduced natural gas supplies, and extreme fossil fuel price volatility—introduced significant uncertainty into the market. These disruptions increased price volatility and likely introduced confounding factors, weakening or obscuring the typical relationship between solar generation and electricity prices. In this context, even periods of high solar output did not consistently lead to lower electricity prices, indicating a potential shift in the underlying data-generating process. This illustrates a form of concept drift, where the statistical relationship between predictors and the response variable changes over time. Detecting such shifts is critical to maintaining model accuracy and interpretability, particularly in high-stakes environments like electricity markets.

This case study therefore furnishes a realistic and challenging testbed for evaluating our adaptive sampling framework. By allocating more labeling resources to regions with high residuals, the method can detect shifts in the predictive relationship such as breakdowns in the expected impact of solar generation on prices, thereby enabling timely responses to changing market conditions.

In many real-world electricity market applications, complete datasets are not freely available. Market operators and data providers frequently sell access to granular, high-frequency data through subscription services, which can make continuous monitoring across all settlement periods (SPs) prohibitively expensive. As a result, practitioners often face data acquisition constraints, having to decide which observations to purchase and analyze. While our case study uses publicly available, country-level UK market data, these constraints become even more relevant when analyzing more plant-specific or locally granular information, where access is typically restricted or costly. In such settings, targeted sampling strategies like the adaptive framework we propose can help reduce data acquisition costs while still preserving predictive performance. This aligns with recent work on regression markets, where selective purchasing of observations has been shown to improve cost–benefit trade-offs in energy forecasting applications \citep{pinson2022regression,goncalves2020towards}.

To evaluate the practical effectiveness of our proposed adaptive sampling strategy, we compared three scenarios:

\begin{enumerate}
    \item Random sampling: in this case, the 8 SPs are selected using a uniform distribution.
    \item \ac{acronym} (with $ \epsilon = 0.5 $): due to practical constraints, we assume we can only monitor prices in 8 of the 48 daily SPs.
    \item Full sampling: all of the 48 daily SPs are observed, providing a full monitoring benchmark for comparison.
\end{enumerate}

In all scenarios, we applied a one-sided EWMA control chart to the top-$r$ average of absolute residuals, $A_t^{(r)}$, where $r$ equals one half of the available daily monitoring budget (i.e., $r=4$ for Random and \ac{acronym}, and $r=24$ for Full sampling). Since establishing analytical control limits (such as the upper control limit, UCL) was impractical due to limited baseline data, we empirically determined the UCL by conducting 1,000 bootstrap simulations on 2020 baseline data using an identical sampling strategy.
Specifically, we set the UCL as the 99.5\% quantile of these simulated statistics. Additionally, as reduced electricity prices typically pose minimal economic risk or operational disruptions, we omitted the lower control limit (LCL), focusing solely on the detection of price surges indicative of potential crises.

\begin{figure}[!htb]
    \centering
    \includegraphics[width=\textwidth]{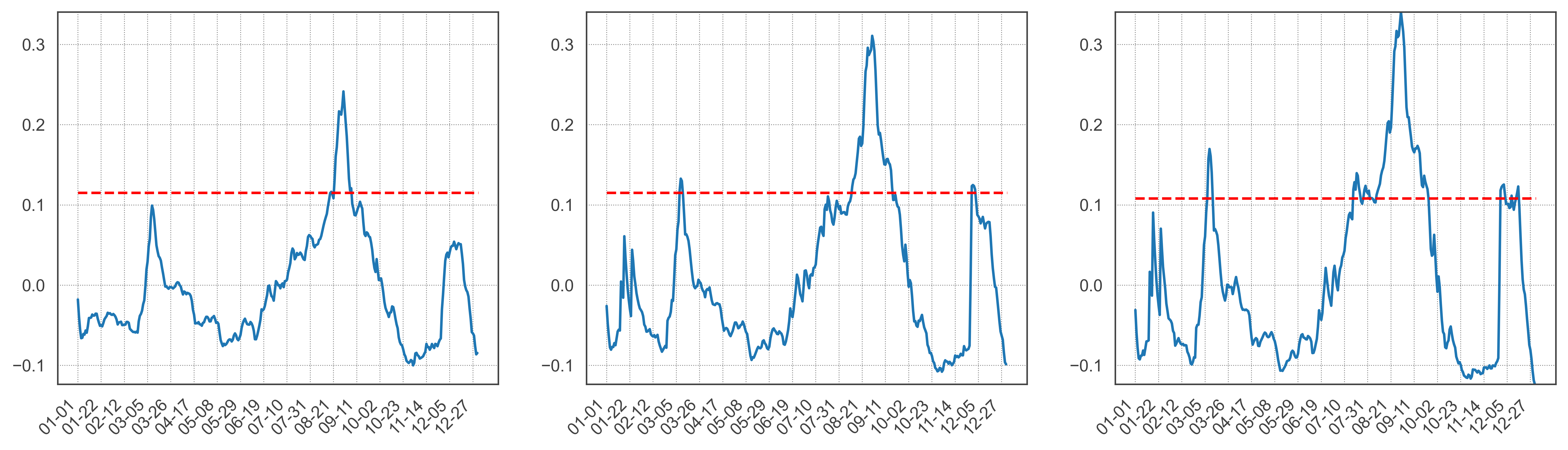}
    \caption{EWMA Control Chart for Detecting Concept Drift in UK Electricity Prices under Random Sampling (left), \ac{acronym} (center), and Full Sampling (right).}
    \label{fig:case_study_result}
\end{figure}

Figure~\ref{fig:case_study_result} illustrates the effectiveness of \ac{acronym} in detecting concept drift within the UK electricity market data. Even though the adaptive strategy was constrained to monitor only eight observations per day, the results using an exploration-exploitation balance parameter $\epsilon=0.5$ (i.e., evenly balancing accept--reject sampling for exploration and inverse transform sampling for exploitation) were remarkably consistent with those obtained from monitoring all available 48 half-hourly observations each day. This observation highlights the efficiency and practicality of our adaptive sampling approach under limited sampling budgets. In contrast, random sampling failed to identify the earlier drift around March 9, detecting only the later shift in August. These results highlight the practical benefits of adaptive sampling in time-sensitive, resource-constrained environments.

\begin{figure}[!htb]
    \centering
    \includegraphics[width=\textwidth]{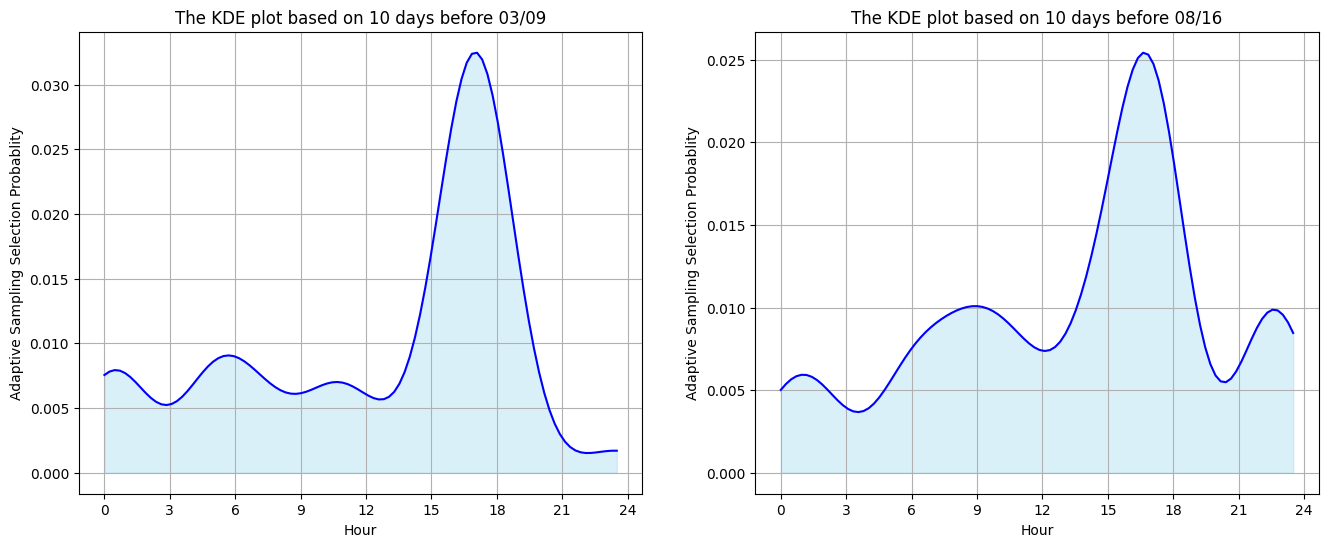}
    \caption{Kernel Density Estimation (KDE) plots of sampling histories for 10 days preceding concept drift detection dates (March 9 and August 16, 2022). The (scaled) day-ahead spot price is modeled as a function of the hour of the day and the predicted solar power penetration using a locally weighted polynomial regression model.}
    \label{fig:kde_case_study}
\end{figure}

To further analyze the detected concept drift, we examined the historical sampling data collected through adaptive sampling. Figure~\ref{fig:kde_case_study} presents Kernel Density Estimation (KDE) plots based on data collected in the 10 days prior to each drift detection event (March 9 and August 16, 2022). These KDE plots clearly reveal that significant concept drift occurred primarily during the hours of approximately 16:00--17:00, where electricity prices sharply increased compared to historical patterns from 2020. To verify these findings, we directly compared electricity price profiles between the stable year (2020) and the crisis year (2022), as depicted in Figure~\ref{fig:lwpr_solar}. Consistent with the KDE results, the most pronounced deviation between the two years occurred around 17:00, with electricity prices in 2022 experiencing a substantial surge. This observation confirms the presence of concept drift and demonstrates the effectiveness of adaptive sampling in promptly identifying critical changes in market behavior.

This case study illustrates that the proposed adaptive sampling approach achieves nearly the same detection performance as exhaustive sampling, while also providing clear insights into precisely when and how concept drift occurs. Consequently, adaptive sampling presents a practical and efficient solution for real-time monitoring and early detection of significant shifts in complex, high-dimensional market systems such as electricity markets.

\section{Conclusions}
\label{sec:conc}

This paper introduced a novel adaptive sampling framework for detecting localized concept drift in regression models under label scarcity. By integrating residual-informed exploration and exploitation strategies with EWMA-based monitoring, our method facilitates efficient drift detection while substantially reducing labeling costs. The exploration strategy, based on accept--reject sampling, ensures effective coverage of underexplored regions, while the exploitation strategy, implemented via inverse transform sampling, focuses resources on areas with high residual uncertainty. Simulation studies across various functions demonstrated that the proposed method consistently outperforms benchmark approaches, including random and Score vector-based sampling, across a wide range of drift magnitudes, region sizes, and data dimensions. Additionally, the framework proved robust under both abrupt and incremental drift scenarios. Our case study on the UK electricity market further confirmed the method’s practical relevance. Despite operating under a limited sampling budget, the adaptive strategy achieved detection performance comparable to full sampling, successfully identifying regime shifts associated with the 2022 energy crisis. These findings underscore the potential of adaptive sampling as a powerful tool for real-time monitoring in complex, dynamic environments where labeling is expensive or constrained.

Future research could extend this work in several directions, such as adaptive methods for tuning the exploration–exploitation balance and theoretical analyses of detection delays and false alarm rates. Furthermore, adapting the framework for multivariate targets and online retraining schemes could enhance its applicability in broader industrial and forecasting settings.

\appendix
\makeatletter
\renewcommand\thesection{Appendix \Alph{section}}
\renewcommand\thesubsection{Appendix \Alph{section}.\arabic{subsection}}
\makeatother

\section{Proof of Proposition 1}
\label{app:proof_exploitation}
\begin{proof}
Because $\mathrm{int}(\mathcal{R})\neq\varnothing$, there exists a nonempty open set $U\subseteq\mathcal{R}$. 
The Gaussian density is strictly positive and continuous on $\mathbb{R}^d$, hence
\[
\mathbb{P}\big(\tilde{\mathbf{x}}\in U\big)\;=\;\int_{U}\phi_h(\mathbf{u}-\mathbf{x})\,d\mathbf{u}\;>\;0,
\]
where $\phi_h$ is the $\mathcal{N}(\mathbf{0},h^2 I_d)$ density. 
Therefore $\mathbb{P}(\tilde{\mathbf{x}}\in\mathcal{R})\ge \mathbb{P}(\tilde{\mathbf{x}}\in U)>0$.

\noindent\textit{Lower bound probability.}
By the law of total probability under the hierarchical scheme,
\[
\mathbb{P}(\tilde{\mathbf{x}}\in\mathcal{R})
= \sum_{i=1}^n \pi_i\,\mathbb{P}(\tilde{\mathbf{x}}\in\mathcal{R}\mid I=i).
\]
Conditional on $I=i$, define the Euclidean closed ball $B_2(\mathbf{x}_i,r_i):=\{\mathbf{u}:\|\mathbf{u}-\mathbf{x}_i\|_2\le r_i\}$ and write $\tilde{\mathbf{x}}=\mathbf{x}_i+h\mathbf{Z}$ where $\mathbf{Z}\sim\mathcal{N}(\mathbf{0},I_d)$. Then, we obtain the event inclusions 
\[
\big\{\|\mathbf{Z}\|_2\le r_i/h\big\}
\ \Longleftrightarrow\
\big\{\|\tilde{\mathbf{x}}-\mathbf{x}_i\|_2\le r_i\big\}
\ \subseteq\
\big\{\tilde{\mathbf{x}}\in B_2(\mathbf{x}_i,r_i)\big\}
\ \subseteq\
\big\{\tilde{\mathbf{x}}\in\mathcal{R}\big\}.
\]
Taking probabilities gives
\[
\mathbb{P}(\tilde{\mathbf{x}}\in\mathcal{R}\mid I=i)
\ \ge\
\mathbb{P}\big(\tilde{\mathbf{x}}\in B_2(\mathbf{x}_i,r_i)\mid I=i\big)
\ =\
\mathbb{P}\big(\|\mathbf{Z}\|_2\le r_i/h\big).
\]
Combining with the total probability decomposition yields
\[
\mathbb{P}(\tilde{\mathbf{x}}\in\mathcal{R})
\ \ge\
\sum_{i=1}^n \pi_i\,\mathbb{P}\!\big(\|\mathbf{Z}\|_2\le r_i/h\big),
\]
which is the claimed lower bound.
\end{proof}

\section{Proof of Proposition 2}
\label{app:proof_exploration}

\begin{proof}
Fix $c_0\in\mathcal{G}$ and let $t_0$ denote its last visit time. While $c_0$ remains unvisited after $t_0$, the acceptance probability $p_{c_0}(t)$ is non-decreasing in $t$, and in particular $p_{c_0}(t)=1$ for all $t\ge t_0+|\mathcal{G}|$.

At time $t$, proposals are i.i.d. uniform over $\mathcal{G}$ and the procedure stops once $m_e$ proposals have been accepted. Let $B_t$ be the event that $c_0$ is proposed at least once before stopping at time $t$. Since the first $m_e$ proposals necessarily occur before stopping, we have the lower bound
\[
\mathbb{P}(B_t)\ \ge\ 1-\Big(1-\tfrac{1}{|\mathcal{G}|}\Big)^{m_e}\ =:\ \alpha\in(0,1].
\]
and the events $\{B_t\}_{t\ge1}$ are independent across $t$. Hence
\[
\sum_{t=t_0+|\mathcal{G}|}^{\infty}\mathbb{P}(B_t)\ \ge\ \sum_{t=t_0+|\mathcal{G}|}^{\infty}\alpha\ =\ \infty.
\]
By the second Borel--Cantelli lemma, $B_t$ occurs infinitely often almost surely. For every such $t\ge t_0+|\mathcal{G}|$ with $B_t$ occurring, the first proposal of $c_0$ at time $t$ is accepted since $p_{c_0}(t)=1$, so $c_0$ is visited infinitely often almost surely. Therefore $\mathbb{P}(c_0\in\mathcal{U})=0$ for each $c_0$, and because $\mathcal{G}$ is finite we conclude $\mathbb{P}(\mathcal{U}=\varnothing)=1$.
\end{proof}

\if0\blind
{
    \section*{Disclosure of Interest}
    The authors have no conflicts of interest to disclose. 

    \section*{Funding}
    No funding was received for this work.
} \fi

\section*{Data Availability Statement}
The simulation datasets used in this study are generated from test functions fully specified in the manuscript.

The datasets used in the case study were derived from publicly available UK electricity market data, including day-ahead electricity prices from the APX and aggregated solar generation statistics. These data are in the public domain and can be accessed through official sources such as the UK National Energy System Operator (NESO) Data Portal (https://www.neso.energy/data-portal) and Elexon (https://bmrs.elexon.co.uk/market-index-prices). The data were scaled and aggregated for the purposes of this study, and derived datasets generated during the current study are available from the corresponding author upon reasonable request.

\bibliographystyle{chicago}
\bibliography{references}

\begin{thebibliography}{}

\bibitem[\protect\citeauthoryear{Baier, Schlör, Schöffer, and Kühl}{Baier et~al.}{2021}]{Baier2021}
Baier, L., T.~Schlör, J.~Schöffer, and N.~Kühl (2021, 7).
\newblock Detecting concept drift with neural network model uncertainty.
\newblock {\em Hawaii International Conference on System Sciences (HICSS) 2023\/}.

\bibitem[\protect\citeauthoryear{Cacciarelli and Kulahci}{Cacciarelli and Kulahci}{2024}]{OALSurvey}
Cacciarelli, D. and M.~Kulahci (2024).
\newblock Active learning for data streams: a survey.
\newblock {\em Machine Learning\/}~{\em 113\/}(1), 185--239.

\bibitem[\protect\citeauthoryear{Crowder and Hamilton}{Crowder and Hamilton}{1992}]{crowder1992ewma}
Crowder, S.~V. and M.~D. Hamilton (1992).
\newblock An ewma for monitoring a process standard deviation.
\newblock {\em Journal of Quality Technology\/}~{\em 24\/}(1), 12--21.

\bibitem[\protect\citeauthoryear{Devroye}{Devroye}{1986}]{devroye1986sample}
Devroye, L. (1986).
\newblock Sample-based non-uniform random variate generation.
\newblock In {\em Proceedings of the 18th conference on Winter simulation}, pp.\  260--265.

\bibitem[\protect\citeauthoryear{Duong-Tran, Dastoorian, and Wells}{Duong-Tran et~al.}{2022}]{duong2022revisiting}
Duong-Tran, D., R.~Dastoorian, and L.~Wells (2022).
\newblock Revisiting the one-sided ewma control chart.
\newblock {\em Journal of applied research on industrial engineering\/}~{\em 9\/}(2), 151--164.

\bibitem[\protect\citeauthoryear{Estrada~Gómez, Li, and Paynabar}{Estrada~Gómez et~al.}{2022}]{estradagomez2022adaptive}
Estrada~Gómez, A.~M., D.~Li, and K.~Paynabar (2022).
\newblock An adaptive sampling strategy for online monitoring and diagnosis of high-dimensional streaming data.
\newblock {\em Technometrics\/}~{\em 64\/}(2), 253--269.

\bibitem[\protect\citeauthoryear{Fabra}{Fabra}{2023}]{fabra2023reforming}
Fabra, N. (2023).
\newblock Reforming european electricity markets: Lessons from the energy crisis.
\newblock {\em Energy economics\/}~{\em 126}, 106963.

\bibitem[\protect\citeauthoryear{Friedman, Grosse, and Stuetzle}{Friedman et~al.}{1983}]{friedman1983multidimensional}
Friedman, J.~H., E.~Grosse, and W.~Stuetzle (1983).
\newblock Multidimensional additive spline approximation.
\newblock {\em SIAM Journal on Scientific and Statistical Computing\/}~{\em 4\/}(2), 291--301.

\bibitem[\protect\citeauthoryear{Gama, Medas, Castillo, and Rodrigues}{Gama et~al.}{2004}]{Gama2004}
Gama, J., P.~Medas, G.~Castillo, and P.~Rodrigues (2004).
\newblock Learning with drift detection.
\newblock {\em Lecture Notes in Computer Science (including subseries Lecture Notes in Artificial Intelligence and Lecture Notes in Bioinformatics)\/}~{\em 3171}, 286--295.

\bibitem[\protect\citeauthoryear{Gama, \v{Z}liobaitundefined, Bifet, Pechenizkiy, and Bouchachia}{Gama et~al.}{2014}]{Gama2014}
Gama, J.~a., I.~\v{Z}liobaitundefined, A.~Bifet, M.~Pechenizkiy, and A.~Bouchachia (2014, March).
\newblock A survey on concept drift adaptation.
\newblock {\em ACM Comput. Surv.\/}~{\em 46\/}(4), 1--37.

\bibitem[\protect\citeauthoryear{Gan}{Gan}{1995}]{gan1995joint}
Gan, F. (1995).
\newblock Joint monitoring of process mean and variance using exponentially weighted moving average control charts.
\newblock {\em Technometrics\/}~{\em 37\/}(4), 446--453.

\bibitem[\protect\citeauthoryear{Goncalves, Pinson, and Bessa}{Goncalves et~al.}{2020}]{goncalves2020towards}
Goncalves, C., P.~Pinson, and R.~J. Bessa (2020).
\newblock Towards data markets in renewable energy forecasting.
\newblock {\em IEEE Transactions on Sustainable Energy\/}~{\em 12\/}(1), 533--542.

\bibitem[\protect\citeauthoryear{IEA}{IEA}{2023}]{IEA2023}
IEA (2023, September).
\newblock Net zero roadmap: A global pathway to keep the 1.5 °c goal in reach -- 2023 update.
\newblock Technical report, International Energy Agency.

\bibitem[\protect\citeauthoryear{Imberg, Jonasson, and Axelson-Fisk}{Imberg et~al.}{2020}]{imberg2020optimal}
Imberg, H., J.~Jonasson, and M.~Axelson-Fisk (2020).
\newblock Optimal sampling in unbiased active learning.
\newblock In {\em International Conference on Artificial Intelligence and Statistics}, pp.\  559--569. PMLR.

\bibitem[\protect\citeauthoryear{Ishigami and Homma}{Ishigami and Homma}{1990}]{ishigami1990importance}
Ishigami, T. and T.~Homma (1990).
\newblock An importance quantification technique in uncertainty analysis for computer models.
\newblock In {\em [1990] Proceedings. First international symposium on uncertainty modeling and analysis}, pp.\  398--403. IEEE.

\bibitem[\protect\citeauthoryear{Johnson, Kotz, and Balakrishnan}{Johnson et~al.}{1995}]{johnson1995continuous}
Johnson, N.~L., S.~Kotz, and N.~Balakrishnan (1995).
\newblock {\em Continuous univariate distributions, volume 2}, Volume~2.
\newblock John wiley \& sons.

\bibitem[\protect\citeauthoryear{Krawczyk}{Krawczyk}{2017}]{Krawczyk2017}
Krawczyk, B. (2017, 12).
\newblock Active and adaptive ensemble learning for online activity recognition from data streams.
\newblock {\em Knowledge-Based Systems\/}~{\em 138}, 69--78.

\bibitem[\protect\citeauthoryear{Krempl, {\v{Z}}liobaite, Brzezi{\'n}ski, H{\"u}llermeier, Last, Lemaire, Noack, Shaker, Sievi, Spiliopoulou, et~al.}{Krempl et~al.}{2014}]{krempl2014open}
Krempl, G., I.~{\v{Z}}liobaite, D.~Brzezi{\'n}ski, E.~H{\"u}llermeier, M.~Last, V.~Lemaire, T.~Noack, A.~Shaker, S.~Sievi, M.~Spiliopoulou, et~al. (2014).
\newblock Open challenges for data stream mining research.
\newblock {\em ACM SIGKDD explorations newsletter\/}~{\em 16\/}(1), 1--10.

\bibitem[\protect\citeauthoryear{Kulldorff}{Kulldorff}{1997}]{kulldorff1997spatial}
Kulldorff, M. (1997).
\newblock A spatial scan statistic.
\newblock {\em Communications in Statistics-Theory and methods\/}~{\em 26\/}(6), 1481--1496.

\bibitem[\protect\citeauthoryear{Linkletter, Bingham, Hengartner, Higdon, and Ye}{Linkletter et~al.}{2006}]{linkletter2006variable}
Linkletter, C., D.~Bingham, N.~Hengartner, D.~Higdon, and K.~Q. Ye (2006).
\newblock Variable selection for gaussian process models in computer experiments.
\newblock {\em Technometrics\/}~{\em 48\/}(4), 478--490.

\bibitem[\protect\citeauthoryear{Liu, Mei, and Shi}{Liu et~al.}{2015}]{liu2015adaptive}
Liu, K., Y.~Mei, and J.~Shi (2015).
\newblock An adaptive sampling strategy for online high-dimensional process monitoring.
\newblock {\em Technometrics\/}~{\em 57\/}(3), 305--319.

\bibitem[\protect\citeauthoryear{Liu, Xue, Wu, Zhou, Yang, Li, and Cao}{Liu et~al.}{2021}]{Liu2021}
Liu, S., S.~Xue, J.~Wu, C.~Zhou, J.~Yang, Z.~Li, and J.~Cao (2021).
\newblock Online active learning for drifting data streams.
\newblock {\em IEEE Transactions on Neural Networks and Learning Systems\/}~{\em 34\/}(1), 186--200.

\bibitem[\protect\citeauthoryear{Lu, Liu, Dong, Gu, Gama, and Zhang}{Lu et~al.}{2018}]{Lu2018}
Lu, J., A.~Liu, F.~Dong, F.~Gu, J.~Gama, and G.~Zhang (2018).
\newblock Learning under concept drift: A review.
\newblock {\em IEEE Transactions on Knowledge and Data Engineering\/}~{\em 31\/}(12), 2346--2363.

\bibitem[\protect\citeauthoryear{Lucas and Saccucci}{Lucas and Saccucci}{1990}]{lucas1990exponentially}
Lucas, J.~M. and M.~S. Saccucci (1990).
\newblock Exponentially weighted moving average control schemes: properties and enhancements.
\newblock {\em Technometrics\/}~{\em 32\/}(1), 1--12.

\bibitem[\protect\citeauthoryear{Mei}{Mei}{2011}]{mei2011quickest}
Mei, Y. (2011).
\newblock Quickest detection in censoring sensor networks.
\newblock In {\em 2011 IEEE International Symposium on Information Theory Proceedings}, pp.\  2148--2152. IEEE.

\bibitem[\protect\citeauthoryear{Mohamad, Bouchachia, and Sayed-Mouchaweh}{Mohamad et~al.}{2018}]{Mohamad2018}
Mohamad, S., A.~Bouchachia, and M.~Sayed-Mouchaweh (2018, 1).
\newblock A bi-criteria active learning algorithm for dynamic data streams.
\newblock {\em IEEE Transactions on Neural Networks and Learning Systems\/}~{\em 29}, 74--86.

\bibitem[\protect\citeauthoryear{Montgomery}{Montgomery}{2020}]{montgomery2020introduction}
Montgomery, D.~C. (2020).
\newblock {\em Introduction to statistical quality control}.
\newblock John wiley \& sons.

\bibitem[\protect\citeauthoryear{Morales, Conejo, Madsen, Pinson, and Zugno}{Morales et~al.}{2013}]{morales2013integrating}
Morales, J.~M., A.~J. Conejo, H.~Madsen, P.~Pinson, and M.~Zugno (2013).
\newblock {\em Integrating renewables in electricity markets: operational problems}, Volume 205.
\newblock Springer Science \& Business Media.

\bibitem[\protect\citeauthoryear{Nabhan, Mei, and Shi}{Nabhan et~al.}{2021}]{nabhan2021correlation}
Nabhan, M., Y.~Mei, and J.~Shi (2021).
\newblock Correlation-based dynamic sampling for online high dimensional process monitoring.
\newblock {\em Journal of Quality Technology\/}~{\em 53\/}(3), 289--308.

\bibitem[\protect\citeauthoryear{Pinson, Han, and Kazempour}{Pinson et~al.}{2022}]{pinson2022regression}
Pinson, P., L.~Han, and J.~Kazempour (2022).
\newblock Regression markets and application to energy forecasting.
\newblock {\em Top\/}~{\em 30\/}(3), 533--573.

\bibitem[\protect\citeauthoryear{Reisi~Gahrooei, Paynabar, Pacella, and Colosimo}{Reisi~Gahrooei et~al.}{2019}]{reisi2019adaptive}
Reisi~Gahrooei, M., K.~Paynabar, M.~Pacella, and B.~M. Colosimo (2019).
\newblock An adaptive fused sampling approach of high-accuracy data in the presence of low-accuracy data.
\newblock {\em IISE Transactions\/}~{\em 51\/}(11), 1251--1264.

\bibitem[\protect\citeauthoryear{Richter, Shi, Chen, Rahnenf{\"u}hrer, and Lang}{Richter et~al.}{2020}]{richter2020model}
Richter, J., J.~Shi, J.-J. Chen, J.~Rahnenf{\"u}hrer, and M.~Lang (2020).
\newblock Model-based optimization with concept drifts.
\newblock In {\em Proceedings of the 2020 genetic and evolutionary computation conference}, pp.\  877--885.

\bibitem[\protect\citeauthoryear{Robert, Casella, and Casella}{Robert et~al.}{1999}]{robert1999monte}
Robert, C.~P., G.~Casella, and G.~Casella (1999).
\newblock {\em Monte Carlo statistical methods}, Volume~2.
\newblock Springer.

\bibitem[\protect\citeauthoryear{Rubinstein and Kroese}{Rubinstein and Kroese}{2016}]{rubinstein2016simulation}
Rubinstein, R.~Y. and D.~P. Kroese (2016).
\newblock {\em Simulation and the Monte Carlo method}.
\newblock John Wiley \& Sons.

\bibitem[\protect\citeauthoryear{Shan, Zhang, Liu, and Liu}{Shan et~al.}{2019}]{Shan2019}
Shan, J., H.~Zhang, W.~Liu, and Q.~Liu (2019, 2).
\newblock Online active learning ensemble framework for drifted data streams.
\newblock {\em IEEE Transactions on Neural Networks and Learning Systems\/}~{\em 30}, 486--498.

\bibitem[\protect\citeauthoryear{Silverman}{Silverman}{2018}]{silverman2018density}
Silverman, B.~W. (2018).
\newblock {\em Density estimation for statistics and data analysis}.
\newblock Routledge.

\bibitem[\protect\citeauthoryear{Soares and Araújo}{Soares and Araújo}{2015}]{Soares2015}
Soares, S.~G. and R.~Araújo (2015, 4).
\newblock A dynamic and on-line ensemble regression for changing environments.
\newblock {\em Expert Systems with Applications\/}~{\em 42}, 2935--2948.

\bibitem[\protect\citeauthoryear{Sun, Tang, Zhu, and Yao}{Sun et~al.}{2018}]{sun2018concept}
Sun, Y., K.~Tang, Z.~Zhu, and X.~Yao (2018).
\newblock Concept drift adaptation by exploiting historical knowledge.
\newblock {\em IEEE transactions on neural networks and learning systems\/}~{\em 29\/}(10), 4822--4832.

\bibitem[\protect\citeauthoryear{Sutton, Barto, et~al.}{Sutton et~al.}{1998}]{sutton1998reinforcement}
Sutton, R.~S., A.~G. Barto, et~al. (1998).
\newblock {\em Reinforcement learning: An introduction}.
\newblock MIT press Cambridge.

\bibitem[\protect\citeauthoryear{Suárez-Cetrulo, Quintana, and Cervantes}{Suárez-Cetrulo et~al.}{2023}]{quintana}
Suárez-Cetrulo, A.~L., D.~Quintana, and A.~Cervantes (2023, 3).
\newblock A survey on machine learning for recurring concept drifting data streams.
\newblock {\em Expert Systems with Applications\/}~{\em 213}, 118934.

\bibitem[\protect\citeauthoryear{Tango and Takahashi}{Tango and Takahashi}{2012}]{tango2012flexible}
Tango, T. and K.~Takahashi (2012).
\newblock A flexible spatial scan statistic with a restricted likelihood ratio for detecting disease clusters.
\newblock {\em Statistics in medicine\/}~{\em 31\/}(30), 4207--4218.

\bibitem[\protect\citeauthoryear{Xian, Zhang, Bonk, and Liu}{Xian et~al.}{2021}]{xian2021rank}
Xian, X., C.~Zhang, S.~Bonk, and K.~Liu (2021).
\newblock Online monitoring of big data streams: A rank-based sampling algorithm by data augmentation.
\newblock {\em Journal of Quality Technology\/}~{\em 53\/}(2), 135--153.

\bibitem[\protect\citeauthoryear{Zan, Wang, and Xian}{Zan et~al.}{2023}]{zan2023spatial}
Zan, X., D.~Wang, and X.~Xian (2023).
\newblock Spatial rank-based augmentation for nonparametric online monitoring and adaptive sampling of big data streams.
\newblock {\em Technometrics\/}~{\em 65\/}(2), 243--256.

\bibitem[\protect\citeauthoryear{Zhang, Liu, Shan, and Liu}{Zhang et~al.}{2018}]{Zhang2018}
Zhang, H., W.~Liu, J.~Shan, and Q.~Liu (2018).
\newblock Online active learning paired ensemble for concept drift and class imbalance.
\newblock {\em IEEE Access\/}~{\em 6}, 73815--73828.

\bibitem[\protect\citeauthoryear{Zhang, Bui, and Apley}{Zhang et~al.}{2023}]{zhang2023concept}
Zhang, K., A.~T. Bui, and D.~W. Apley (2023).
\newblock Concept drift monitoring and diagnostics of supervised learning models via score vectors.
\newblock {\em Technometrics\/}~{\em 65\/}(2), 137--149.

\end{thebibliography}

\end{document}